\documentclass[10pt,a4paper]{article}
\usepackage[margin=1in]{geometry}
\usepackage[numbers,sort&compress]{natbib}
\usepackage{authblk}
\usepackage[font=small]{caption}
\usepackage{graphicx}%
\usepackage{multirow}%
\usepackage{amsmath,amssymb,amsfonts}%
\usepackage{amsthm}%
\usepackage{mathrsfs}%
\usepackage[title]{appendix}%
\usepackage{xcolor}%
\usepackage{textcomp}%
\usepackage{booktabs}%
\usepackage{algorithm}%
\usepackage{algorithmicx}%
\usepackage{algpseudocode}%
\usepackage{listings}%
\usepackage{array}%
\usepackage{lscape}%
\usepackage{float}%
\usepackage{pifont}
\usepackage{makecell} 
\usepackage[edges]{forest}
\usepackage{tikz}
\usetikzlibrary{trees, positioning, shapes.geometric, arrows.meta}

\newcommand{\cmark}{\ding{51}}%
\newcommand{\xmark}{\ding{55}}%

\definecolor{mygrey}{RGB}{128,128,128}
\definecolor{hidden-draw}{RGB}{100,100,100}
\definecolor{hidden-pink}{RGB}{255,182,193}

\tikzstyle{my-box}=[
    rectangle,
    draw=hidden-draw,
    rounded corners,
    text opacity=1,
    minimum height=1.6em,
    minimum width=6em,
    inner sep=3pt,
    align=center,
    fill opacity=.5,
    line width=0.8pt,
]

\tikzstyle{leaf}=[my-box,
    minimum height=1.6em,
    fill=hidden-pink!80,
    text=black,
    align=left,
    font=\normalsize,
    inner xsep=3pt,
    inner ysep=5pt,
    line width=0.8pt,
]

\usepackage[colorlinks=true,linkcolor=blue,citecolor=blue,urlcolor=blue]{hyperref}

\title{RISC-V and Machine Learning: A Survey}
\author[1]{Shriman Keshri\thanks{shriman.keshri@niser.ac.in}}
\author[2,$\dagger$]{Apparna Singh\thanks{apparna.s2021btcseai@srisriuniversity.edu.in}}
\author[3,$\dagger$]{Chinmaya Kumar Palo\thanks{chinmayakumarpalo.official@gmail.com}}
\author[2,$\dagger$]{Shreya Adya\thanks{adyashreya2@gmail.com}}
\author[1,4]{Subhankar Mishra\thanks{Corresponding author: smishra@niser.ac.in}}
\affil[1]{National Institute of Science Education and Research, Bhubaneswar, Odisha, India}
\affil[2]{Sri Sri University, Cuttack, Odisha, India}
\affil[3]{Gandhi Institute of Engineering and Technology University, Gunupur, Odisha, India}
\affil[4]{Homi Bhabha National Institute, Mumbai, Maharashtra, India}
\affil[$\dagger$]{Work conducted at NISER}
\date{}

\begin{document}

\maketitle
\begingroup
\renewcommand{\thefootnote}{}%
\footnotetext{This is the authors' accepted manuscript of: S.~Keshri, A.~Singh, C.~K.~Palo, S.~Adya, S.~Mishra, ``RISC-V and machine learning: a survey,'' \emph{The Journal of Supercomputing}, vol.~82, no.~8, art.~424, 2026. Published 19 May 2026. The Version of Record is available at \url{https://doi.org/10.1007/s11227-026-08463-z}.}%
\endgroup

\begin{abstract}
The intersection of open-source processor architectures and machine learning is driving the demand for customizable, efficient, and accessible hardware. This survey examines the state of the RISC-V ISA in machine learning applications, analyzing current capabilities, challenges, and future directions based on recent research. The analysis covers academic and commercial implementations, software frameworks, and real-world applications. The RISC-V machine learning ecosystem is evaluated, from instruction set extensions and core implementations to compiler optimizations and deployment strategies. Key contributions include a unified taxonomy of RISC-V ML implementations, a comparative analysis of performance and design trade-offs, an evaluation of software toolchain maturity, and the identification of emerging trends in instruction set extensions and specialized accelerators. Findings reveal progress in energy efficiency, specialized instruction development, and framework integration, while highlighting challenges in standardization, verification complexity, and ecosystem fragmentation. The analysis proposes four research directions to address current limitations: specialized neural processing extensions, adaptive and modular processor architectures, security frameworks, and energy-efficient multi-domain architectures. These directions provide a roadmap for advancing RISC-V as a foundational platform for next-generation machine learning systems.
\end{abstract}


\section{Introduction}

Proprietary hardware architectures face challenges from increasing demand for open-source, customizable solutions \cite{clifford2024locking}. This shift stems from requirements for low-power, cost-effective hardware accelerators, semiconductor shortages, and interest in open-source hardware. Supply chain limitations became apparent during disruptions like the COVID-19 pandemic and geopolitical tensions \cite{hadfield2023regulatory}, increasing interest in affordable and adaptable hardware solutions \cite{ledin2022modern}.
RISC-V \cite{kanter2016risc} is an open-source instruction set architecture (ISA) that allows unrestricted use, modification, and distribution without licensing fees. This openness supports a global developer community \cite{ince2019building} and benefits small and medium-sized enterprises (SMEs) and startups \cite{di2019case,di2019leveraging,raveendran2016risc}. RISC-V enables customization through specialized instructions and accelerators for specific application domains without vendor lock-in or licensing constraints.

RISC-V is used in machine learning, particularly for edge computing applications including speech recognition, object detection, and text processing. Demand for specialized processing units for edge computations \cite{capra2019edge, flamand2018gap, montesdeoca2023softprocessor} has motivated research into RISC-V's architectural adaptability. The modular nature of RISC-V allows implementation of custom extensions optimized for neural network operations, tensor processing, and other ML computations.
Recent work demonstrates RISC-V's role in machine learning through contemporary core implementations \cite{chander2022soft, kong2019airv, garofalo2022darkside, garofalo2021xpulpnn, hou2020rvtensor, ueyoshi2022diana, muller2024gap9shield} and new developments \cite{marvel2025framework, nunes2025accelerating, colagrande2025zero, wang2025vexp}. Analysis includes performance comparisons, architectural trade-offs, and efficiency metrics across RISC-V implementations for machine learning workloads.
Additional studies spanning modular processors, verification frameworks, side-channel resilience, memory behavior, and comparative system evaluation are cited directly in this text for traceability \cite{vacca2024rempro,schiavone_open-source_2018,rutishauser2024xtern,maras2024extending,peng2024performance,wu2024design,simson2024comparative,kroger2025federated,yang2024hardware,bhade2024lightweight,asano2024simulation,jiang2024pcg,miteloudi2024plan,christensen2021wire,liu2024exploring,liu_performance_2022,yu_case_2024}.

Literature reviews on RISC-V and machine learning are limited. An earlier survey by Nicholas et al. was published in 2020 \cite{nicholas_survey_2020}; subsequent developments motivate an updated review with post-2020 emphasis. Recent RISC-V-focused ML surveys also include the ecosystem survey by Kalapothas et al. \cite{kalapothas2023survey} and the edge deep-learning perspective by Agosta et al. \cite{agosta2025deep}. Existing surveys often lack quantitative comparative analysis and concrete conclusions about different approaches.
Table \ref{table:related-papers} presents related research papers and their scope, showing how this survey extends existing literature. Previous surveys have covered various aspects of RISC-V; this work addresses gaps by providing coverage of recent developments, quantitative performance analysis, and practical implementation guidance. Compared with recent RISC-V-for-ML survey literature \cite{kalapothas2023survey,agosta2025deep}, this survey emphasizes cross-stack linkage from ISA extensions to software frameworks and application-level outcomes. The survey offers systematic evaluation of both hardware implementations and software frameworks.

This survey addresses gaps through analysis of contemporary RISC-V cores and implementations. We evaluate software frameworks and tools that facilitate AI accelerator development on RISC-V. Application evaluations demonstrate how RISC-V implementations address practical machine learning challenges through performance metrics and deployment scenarios.

\begin{table}[h]
\caption{Overview of Related Research Papers and Their Scope}
\label{table:related-papers}
\centering
\begin{tabular}{lcccc}
\toprule
\multicolumn{1}{c}{Paper} & \multicolumn{4}{c}{Scope} \\
\midrule
& RISC-V Extensions & Custom ISA & IoT Security & Systematic Review \\
\midrule
Cui et al. \cite{cui_risc-v_2023} & \cmark & \cmark & \cmark & \cmark \\
Lu et al. \cite{lu_survey_2021} & \xmark & \xmark & \cmark & \cmark \\
Dorflinger et al. \cite{dorflinger_comparative_2021} & \cmark & \cmark & \xmark & \xmark \\
Mezger et al. \cite{mezger_survey_2022} & \cmark & \cmark & \xmark & \cmark \\
Nicholas et al. \cite{nicholas_survey_2020} & \cmark & \cmark & \cmark & \cmark \\
Kalapothas et al. \cite{kalapothas2023survey} & \cmark & \cmark & \xmark & \cmark \\
Agosta et al. \cite{agosta2025deep} & \cmark & \cmark & \xmark & \cmark \\
\textbf{Our Survey} & \cmark & \cmark & \cmark & \cmark \\
\bottomrule
\end{tabular}
\end{table}
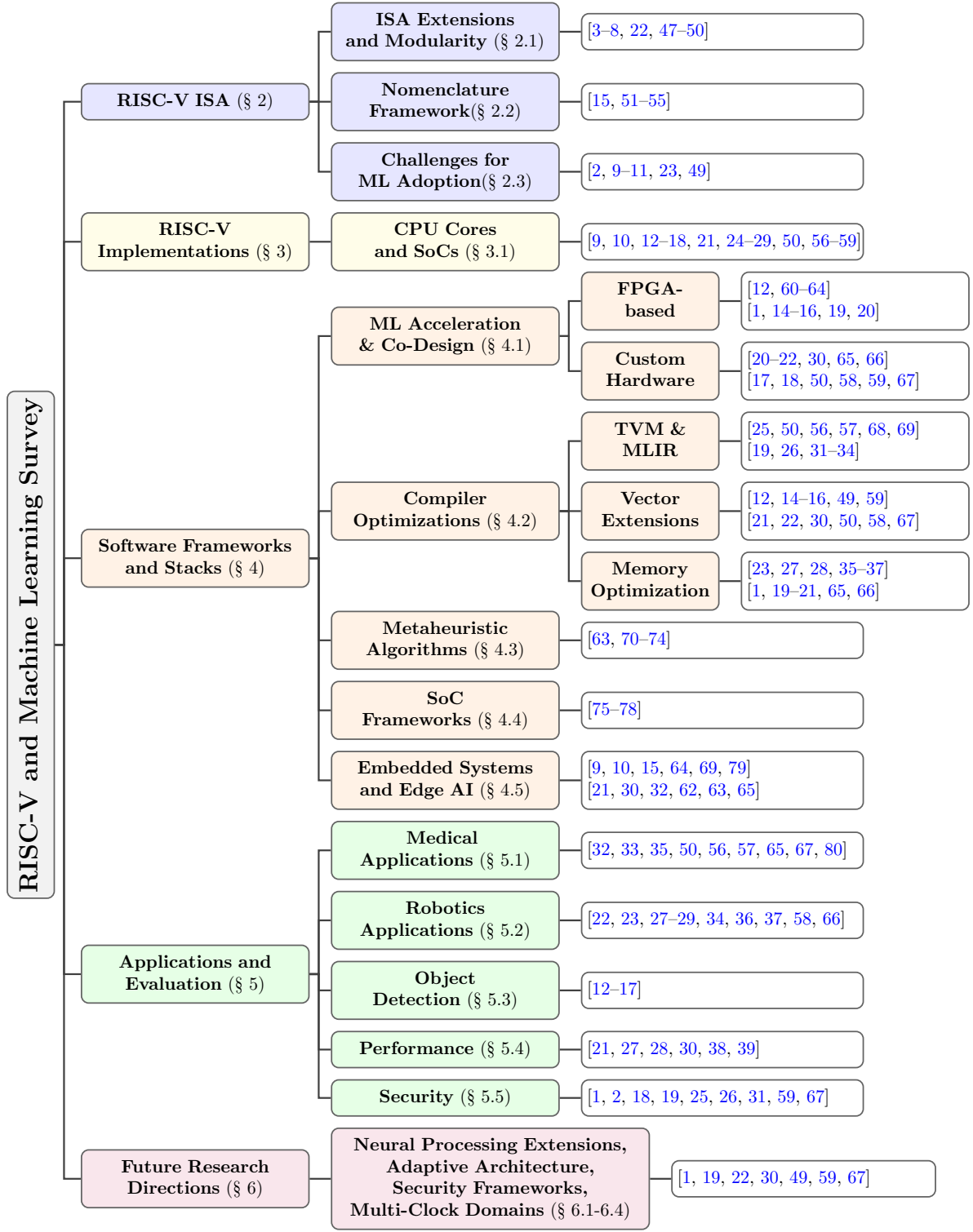
\begin{figure*}[htbp]
    \centering
    \resizebox{0.98\textwidth}{!}{%
        \begin{forest}
            forked edges,
            for tree={
                grow=east,
                reversed=true,
                anchor=base west,
                parent anchor=east,
                child anchor=west,
                base=center,
                font=\Large,
                rectangle,
                draw=hidden-draw,
                rounded corners,
                align=center,
                minimum width=6em,      
                edge+={darkgray, line width=1pt},
                s sep=6pt,              
                inner xsep=3pt,         
                inner ysep=4pt,         
                line width=0.8pt,
                ver/.style={rotate=90, child anchor=north, parent anchor=south, anchor=center},
            },
            where level=1{text width=12em,font=\normalsize, text centered}{},
            where level=2{text width=12em,font=\normalsize, text centered}{},
            where level=3{text width=7em,font=\normalsize}{},
            where level=4{text width=12em,font=\normalsize, align=left}{},
            [
                \textbf{RISC-V and Machine Learning Survey},  fill=mygrey!10, ver
                [
                    \textbf{RISC-V ISA} (\S~2), fill=blue!10
                    [
                        \textbf{ISA Extensions} \\ \textbf{and Modularity} (\S~2.1),
                        fill=blue!10
                        [
                            \cite{kanter2016risc,Waterman2014TheRI,asanovic2016rocket,raveendran2016risc,ince2019building,di2019leveraging,di2019case,ledin2022modern,gupta_challenges_2023,wang2025vexp,bertaccini2024minifloats},
                            align=left, text width=15em
                        ]
                    ]
                    [
                        \textbf{Nomenclature} \\ \textbf{Framework}(\S~2.2), fill=blue!10
                        [
                            \cite{waterman2019risc,embeddev_risc-v_nodate,prakash_cfu_2023,balasubramanian_designing_2024,garofalo2021xpulpnn,lee_accelerating_2023},
                            align=left, text width=15em
                        ]
                    ]
                    [
                        \textbf{Challenges for} \\ \textbf{ML Adoption}(\S~2.3), fill=blue!10
                        [
                            \cite{hadfield2023regulatory,montesdeoca2023softprocessor,flamand2018gap,capra2019edge,vacca2024rempro,gupta_challenges_2023},
                            align=left, text width=15em
                        ]
                    ]
                ]
                [
                    \textbf{RISC-V} \\ \textbf{Implementations} (\S~3), fill=yellow!10
                    [
                        \textbf{CPU Cores} \\ \textbf{and SoCs} (\S~3.1), fill=yellow!10
                        [
                            \cite{schiavone_open-source_2018,flamand2018gap,capra2019edge,kong2019airv,hou2020rvtensor,garofalo2021xpulpnn,chander2022soft,garofalo2022darkside,ueyoshi2022diana,muller2024gap9shield,bertaccini2024minifloats,manoni2024nars,armeniakos2024mixed,rutishauser2024xtern,maras2024extending,peng2024performance,wu2024design,simson2024comparative,colagrande2025zero,rodrigo2025vseek,sabih2025sparse},
                            align=left, text width=15em 
                        ]
                    ]
                ]
                [
                    \textbf{Software Frameworks} \\ \textbf{and Stacks} (\S~4), fill=orange!10
                                    [
                                        \textbf{ML Acceleration} \\ \textbf{\& Co-Design} (\S~4.1), fill=orange!10
                                                                                [
                                                                                    \textbf{FPGA-} \\ \textbf{based}, fill=orange!10, text centered
                                                                                    [
                                                                                        \cite{roorda_fpga_2022,vreca_towards_2023,cheikh_tourad_generic_2022,xiao_gahls_2023,mika_vedliot_2023,chander2022soft} \\ \cite{garofalo2022darkside,garofalo2021xpulpnn,hou2020rvtensor,clifford2024locking,marvel2025framework,nunes2025accelerating},
                                                                                        align=left, 
                                                                                    ]
                                                                                ]
                                                                                [
                                                                                    \textbf{Custom} \\ \textbf{Hardware}, fill=orange!10, text centered
                                                                                    [
                                                                                        \cite{leliwa2025optimised,nunes2025accelerating,colagrande2025zero,vergos2025svm,wang2025vexp,kroger2025federated} \\ \cite{rodrigo2025vseek,sabih2025sparse,daghero2025lightweight,ueyoshi2022diana,muller2024gap9shield,bertaccini2024minifloats},
                                                                                        align=left, 
                                                                                    ]
                                                                                ]
                                                                            ]
                                                                            [
                                                                                \textbf{Compiler} \\ \textbf{Optimizations} (\S~4.2), fill=orange!10
                                                                                [
                                                                                    \textbf{TVM \&} \\ \textbf{MLIR}, fill=orange!10, text centered
                                                                                    [
                                                                                        \cite{bik_compiler_2022,rotem_glow_2019,bertaccini2024minifloats,manoni2024nars,armeniakos2024mixed,rutishauser2024xtern} \\ \cite{maras2024extending,yang2024hardware,bhade2024lightweight,asano2024simulation,jiang2024pcg,marvel2025framework},
                                                                                        align=left, 
                                                                                    ]
                                                                                ]
                                                                                [
                                                                                    \textbf{Vector} \\ \textbf{Extensions}, fill=orange!10, text centered
                                                                                    [
                                                                                        \cite{chander2022soft,garofalo2022darkside,garofalo2021xpulpnn,hou2020rvtensor,gupta_challenges_2023,sabih2025sparse} \\ \cite{wang2025vexp,rodrigo2025vseek,daghero2025lightweight,kroger2025federated,colagrande2025zero,bertaccini2024minifloats},
                                                                                        align=left,
                                                                                    ]
                                                                                ]
                                                                                [
                                                                                    \textbf{Memory} \\ \textbf{Optimization}, fill=orange!10, text centered
                                                                                    [
                                                                                        \cite{miteloudi2024plan,peng2024performance,wu2024design,vacca2024rempro,christensen2021wire,liu2024exploring} \\ \cite{nunes2025accelerating,leliwa2025optimised,marvel2025framework,clifford2024locking,vergos2025svm,colagrande2025zero},
                                                                                        align=left,
                                                                                    ]
                                                                                ]
                                                                            ]
                                                                            [
                                                                                \textbf{Metaheuristic} \\ \textbf{Algorithms} (\S~4.3), fill=orange!10
                                                                                [
                                                                                    \cite{zhou_agile_2024,xu_effective_2022,jia_automatic_2023,agostini_mlir-based_2022,vreca_hardwaresoftware_2024,xiao_gahls_2023},
                                                                                    align=left, text width=15em
                                                                                ]
                                                                            ]
                                                                            [
                                                                                \textbf{SoC} \\ \textbf{Frameworks} (\S~4.4), fill=orange!10
                                                                                [
                                                                                    \cite{kermarrec_litex_2020,ramanathan_case_2022,ma_design_2023,ferres_chisel_2023},
                                                                                    align=left, text width=15em
                                                                                ]
                                                                            ]
                                                                            [
                                                                                \textbf{Embedded Systems} \\ \textbf{and Edge AI} (\S~4.5), fill=orange!10
                                                                                [
                                                                                    \cite{capra2019edge,flamand2018gap,mika_vedliot_2023,garofalo2021xpulpnn,rotem_glow_2019,liu_tinyiree_2022} \\ \cite{bhade2024lightweight,leliwa2025optimised,kroger2025federated,colagrande2025zero,cheikh_tourad_generic_2022,xiao_gahls_2023},
                                                                                    align=left, text width=15em
                                                                                ]
                                                                            ]                           ]     [
                    \textbf{Applications and} \\ \textbf{Evaluation} (\S~5), fill=green!10
                    [
                        \textbf{Medical } \\ \textbf{Applications} (\S~5.1), fill=green!10
                        [
                            \cite{yoo_real-time_2024,bhade2024lightweight,asano2024simulation,armeniakos2024mixed,manoni2024nars,bertaccini2024minifloats,miteloudi2024plan,daghero2025lightweight,leliwa2025optimised},
                            align=left, text width=15em
                        ]
                    ]
                    [
                        \textbf{Robotics} \\ \textbf{Applications} (\S~5.2), fill=green!10
                        [
                            \cite{jiang2024pcg,wu2024design,peng2024performance,simson2024comparative,liu2024exploring,christensen2021wire,vacca2024rempro,wang2025vexp,rodrigo2025vseek,vergos2025svm},
                            align=left, text width=15em
                        ]
                    ]
                    [
                        \textbf{Object} \\ \textbf{Detection} (\S~5.3), fill=green!10
                        [
                            \cite{garofalo2022darkside,garofalo2021xpulpnn,hou2020rvtensor,ueyoshi2022diana,chander2022soft,kong2019airv},
                            align=left, text width=15em
                        ]
                    ]
                    [
                        \textbf{Performance} (\S~5.4), fill=green!10
                        [
                            \cite{liu_performance_2022,yu_case_2024,peng2024performance,wu2024design,kroger2025federated,colagrande2025zero},
                            align=left, text width=15em
                        ]
                    ]
                    [
                        \textbf{Security} (\S~5.5), fill=green!10
                        [
                            \cite{muller2024gap9shield,hadfield2023regulatory,clifford2024locking,marvel2025framework,rutishauser2024xtern,maras2024extending,yang2024hardware,sabih2025sparse,daghero2025lightweight},
                            align=left, text width=15em
                        ]
                    ]
                ]
                [
                    \textbf{Future Research} \\ \textbf{Directions} (\S~6), fill=purple!10
                    [
                        \textbf{Neural Processing Extensions,} \\  \textbf{Adaptive Architecture,} \\  \textbf{Security Frameworks,} \\
                        \textbf{Multi-Clock Domains} (\S~6.1-6.4), text width=17em, fill=purple!10
                        [
                            \cite{marvel2025framework,wang2025vexp,clifford2024locking,gupta_challenges_2023,kroger2025federated,daghero2025lightweight,sabih2025sparse},
                            align=left, text width=14em
                        ]
                    ]
                ]
            ]
        \end{forest}%
    }
    \caption{Comprehensive outline of the RISC-V and Machine Learning survey.}
    \label{fig:survey-outline}
\end{figure*}

\subsection{Search Methodology}

A literature search was conducted using Google Scholar, DBLP, IEEE Xplore, ACM Digital Library, arXiv, and Springer, focusing on developments after \cite{nicholas_survey_2020}. The primary systematic focus is post-2020 work, while foundational pre-2020 studies are retained when needed for background, baselines, and terminology alignment. The following search terms were used:

\begin{itemize}
    \item (RISC V | RISC-V) + machine learning
    \item (RISC V | RISC-V) + deep learning
    \item (RISC V | RISC-V) + neural networks
    \item (RISC V | RISC-V) + AI
    \item (RISC V | RISC-V) + accelerators
    \item (RISC V | RISC-V) + hardware acceleration
    \item (RISC V | RISC-V) + high-performance computing
    \item (RISC V | RISC-V) + AI inference
    \item (RISC V | RISC-V) + ML frameworks
\end{itemize}

Papers were included if they: utilized RISC-V to improve ML efficiency; presented ML models optimized for RISC-V platforms; reported implementation results for ML applications on RISC-V hardware; introduced hardware accelerators or software optimizations for RISC-V ML workflows; or provided quantitative performance analysis of RISC-V systems executing ML workloads.
Papers were excluded if they mentioned RISC-V only peripherally, lacked experimental validation, presented purely theoretical work without implementation details, or failed to provide sufficient technical depth for analysis. Quality assessment evaluated technical rigor, clarity of contribution, performance metrics availability, and reproducibility.

\subsection{Organisation and Contributions}

This survey examines RISC-V's impact on machine learning through six sections. Section 2 presents the RISC-V ISA, covering its extension framework, nomenclature conventions, and challenges for ML adoption. Section 3 analyzes RISC-V implementations across academic and commercial cores with comparative evaluation. Section 4 examines software frameworks and stacks, including hardware-software co-design, compiler optimization, metaheuristic algorithms, SoC frameworks, and embedded systems for edge AI. Section 5 validates these frameworks through applications and evaluation organized by application domains, performance characteristics, and security considerations. Section 6 identifies research directions based on current limitations.  Figure \ref{fig:survey-outline} provides an overview of the survey organization.

This survey makes the following contributions:

\begin{itemize}
    \item A cross-stack analysis connecting RISC-V ISA extensions, processor implementations, software frameworks, and ML application outcomes. Unlike prior work that treats these layers separately, this survey traces how ISA-level design decisions affect ML performance from hardware through to deployment.

    \item Normalized comparison tables for academic and commercial RISC-V cores with consistent architectural metrics and per-entry source citations, consolidating data previously spread across individual publications and vendor documentation.

    \item A taxonomy of ML software frameworks organized by their RISC-V integration approach, with comparative analysis of compilation toolchains targeting RISC-V backends.

    \item A synthesis of quantitative results reported across RISC-V ML implementations in multiple application domains, drawing architectural patterns and trade-offs from the surveyed studies rather than original experiments.

    \item Four research directions based on gaps identified in the surveyed literature.
\end{itemize}

\section{RISC-V ISA}
The RISC-V Instruction Set Architecture (ISA) is an open-standard ISA based on reduced instruction set computer (RISC) principles. Unlike proprietary ISAs, RISC-V is freely available under open-source licenses, enabling unrestricted use in academic research, commercial products, and custom hardware implementations. This openness, combined with its modular extension framework, has driven rapid adoption across embedded systems, high-performance computing, and machine learning acceleration. This section covers the ISA's extension framework, nomenclature conventions, and the key challenges for ML adoption on RISC-V platforms.

\subsection{RISC-V ISA Extensions and Modularity}

The RISC-V Instruction Set Architecture (ISA) represents a shift toward modular computing architectures, distinguishing itself from traditional monolithic instruction set designs through its extensibility framework. While conventional architectures typically implement fixed instruction sets that constrain application-specific optimization, RISC-V establishes a modular foundation that enables targeted architectural enhancement through standardized extension mechanisms. This architectural philosophy facilitates domain-specific optimization while maintaining compatibility across diverse implementation contexts, addressing contemporary challenges of specialized computing requirements without sacrificing interoperability.

The extensibility paradigm operates through a hierarchical framework encompassing both standardized and custom extensions, each serving distinct roles within the RISC-V ecosystem. This approach enables architectural customization that directly addresses application-specific computational patterns, memory access behaviors, and performance requirements. The modular design philosophy extends beyond simple instruction addition, encompassing architectural modifications including specialized register files, custom functional units, and domain-specific memory hierarchies. These capabilities prove valuable for machine learning applications, where computational patterns often deviate from general-purpose computing workloads.

Machine learning workloads demonstrate computational characteristics that differ from traditional general-purpose applications, necessitating specialized architectural support for efficient execution. Neural network inference and training exhibit regular computational patterns dominated by matrix operations, convolution kernels, and element-wise vector operations that benefit from specialized instruction support. The RISC-V extension framework enables development of domain-specific instructions that directly target these computational patterns, resulting in performance improvements and energy efficiency gains compared to software-only implementations on general-purpose architectures.

There are two primary categories of RISC-V extensions:

    \subsubsection{Standard Extensions} 
    Standard extensions comprise predefined instruction sets officially ratified by RISC-V International, targeting well-established requirements and application domains. Standard extensions ensure interoperability and software compatibility across various RISC-V implementations, as detailed in Table \ref{table:standard-extensions}. Representative examples include:
    \begin{itemize}
        \item \textbf{RV32I/RV64I}: These form the foundation of all RISC-V implementations, providing basic integer instructions for 32-bit and 64-bit variants, respectively. They are considered ``frozen,'' meaning that their specifications are finalized and will not change.
        \item \textbf{M (Standard Extension for Integer Multiplication and Division):} This extension expands the base ISA to include instructions for efficient multiplication and division operations.
        \item \textbf{A (Standard Extension for Atomic Instructions):} Crucial for synchronization in multi-core systems, this extension introduces atomic instructions that ensure operations execute as a single, indivisible unit.
        \item \textbf{F (Standard Extension for Single-Precision Floating-Point) and D (Standard Extension for Double-Precision Floating-Point):} These extensions cater to applications requiring floating-point arithmetic, offering instructions for single-precision (32-bit) and double-precision (64-bit) operations, respectively.
        \item \textbf{G (Shorthand for I+M+A+F+D):} A common combination of extensions often found in general-purpose RISC-V implementations.
        \item \textbf{Other Standard Extensions}: Beyond the examples mentioned, RISC-V offers a range of additional standard extensions, including those for quad-precision floating-point (Q), compressed instructions (C), bit manipulation (B), vector operations (V), hypervisor functionality (H), and supervisor-level instructions (S). All standard extensions listed in Table \ref{table:standard-extensions} have been ratified by RISC-V International.
    \end{itemize}

    \begin{table}[t]
    \caption{Standard ISA Extensions: This table shows ML-relevant standard ISA extensions with their symbols, descriptions, version numbers, and current ratification status (as of December 2024).}
    \label{table:standard-extensions}
    \centering
    \begin{tabular}{llll}
    \toprule
         \textbf{Name} & \textbf{Description} & \textbf{Version} & \textbf{Status} \\
    \midrule
         M & Integer Multiplication and Division & v2.0 & Ratified \\
         A & Atomic Instructions & v2.1 & Ratified \\
         F & Single-Precision Floating-Point & v2.2 & Ratified \\
         D & Double-Precision Floating-Point & v2.2 & Ratified \\
         G & Shorthand for IMAFD (not a separate extension) & - & - \\
         Q & Quad-Precision Floating-Point & v2.2 & Ratified \\
         C & Compressed Instructions & v2.0 & Ratified \\
         B & Bit Manipulation & v1.0 & Ratified \\
         V & Vector Operations & v1.0 & Ratified \\
         H & Hypervisor & v1.0 & Ratified \\
         S & Supervisor-level Instructions & v1.12 & Ratified \\
    \bottomrule
    \end{tabular}
\end{table}

\begin{table}
\caption{Custom ISA Extensions: This table shows the various custom ISA extensions of the RISC-V processor with their symbol/name, description, and the year in which they were introduced or if they are under development or not. Rows marked with * indicate features that are still under development.}
\label{table:custom-extensions}
\centering
\begin{tabular}{>{\raggedright\arraybackslash}p{0.2\linewidth}>{\raggedright\arraybackslash}p{0.6\linewidth}c}
\toprule
\textbf{Extension} & \textbf{Description} & \textbf{Year} \\
\midrule
Xpulpnn, Xvec & Custom Neural Network and Vector Operations (standard vector is V/RVV) & 2019-2021 \\
K & Cryptography Instructions & 2017 \\
Zicsr & Control and Status Register Support & 2018 \\
Zifencei & Load--store fence instructions & 2018 \\
Zfh & Half-Precision Floating-Point Support (IEEE 754-2008 binary16) & 2021 \\
Smcdeleg, Ssccfg & Supervisor Counter Delegation (M-mode to S-mode counter access) & 2024 * \\
Smstateen & State Enabled Extension & 2023 \\
Zca, Zcb, Zcd, Zce, Zcf, Zcmp, and Zcmt & Code Size Reduction & 2023 \\
Smepmp & PMP Enhancements for memory access and execution prevention in Machine mode & 2021 \\
Zba, Zbb, Zbc, and Zbs & Bit-Manipulation ISA-extensions & 2021 \\
RV32E/RV64E & RV32E and RV64E Base Integer Instruction Sets & 2023 \\
Zfa & Standard Extension for Additional Floating-Point Instructions & 2023 \\
Zvfh, Zvfhmin & Vector Extension for Half-Precision Floating-Point Arithmetic/Vector Extension for Minimal Half-Precision Floating-Point Arithmetic & 2023 \\
Zihintpause & Pause Hint & 2021 \\
Sscofpmf & Count Overflow and Mode-Based Filtering Extension & 2021 \\
Zabha & Byte and Halfword Atomic Memory Operations (8-bit and 16-bit atomics for quantized neural networks) & 2024 \\
\bottomrule
\end{tabular}
\end{table}

    \subsubsection{Custom Extensions} 
    
The extensible nature of the RISC-V ISA enables designers to develop application-specific instructions that provide fine-grained optimization for specialized computational tasks. This architectural flexibility has proven particularly valuable in domains such as machine learning, signal processing, and cryptography, where domain-specific instructions can yield substantial performance improvements. The custom extension ecosystem encompasses several key categories, as outlined in Table \ref{table:custom-extensions}.

Vector processing capabilities are enhanced through custom vector extensions (Xvec) \cite{nunes2025accelerating}, which introduce application-specific vector instruction sets that accelerate computations in signal processing, linear algebra, and scientific simulation applications. These custom extensions complement the standard RISC-V Vector extension (V/RVV) by providing domain-specific optimizations beyond the ratified specification. This extension enables Single Instruction, Multiple Data (SIMD) operations that are fundamental to modern parallel computing workloads \cite{gupta_challenges_2023}. For security-critical applications, the K (Cryptography Instructions) extension provides dedicated hardware-accelerated cryptographic operations, substantially improving both the efficiency and security robustness of encryption and decryption processes. Additionally, the Zfh (Half-Precision Floating-Point) extension, introduced in 2021, provides hardware support for IEEE 754 binary16 half-precision floating-point operations, enabling efficient low-precision arithmetic for ML inference and scientific computing \cite{bertaccini2024minifloats}. The broader custom extension landscape demonstrates RISC-V's remarkable adaptability to emerging computational paradigms and specialized application requirements \cite{armeniakos2024mixed,manoni2024nars}.
Machine learning workloads have driven the development of several specialized RISC-V ISA extensions that address the unique computational and architectural requirements of neural network inference and training \cite{daghero2025lightweight,sabih2025sparse}. The Smcdeleg and Ssccfg (Supervisor Counter Delegation) extensions provide fine-grained performance monitoring capabilities specifically tailored for ML applications, enabling hypervisors to delegate counter access privileges to guest operating systems. This functionality proves crucial for comprehensive neural network performance profiling and dynamic resource allocation optimization in multi-tenant machine learning environments.

Atomic memory operations for quantized neural networks are enhanced through the Zabha (Byte and Halfword Atomic Memory Operations) extension, which supports efficient synchronization primitives for 8-bit and 16-bit data types commonly employed in quantized deep neural networks \cite{chander2022soft}. These atomic operations significantly reduce memory contention in multi-core ML inference scenarios, enabling more efficient parallel processing of quantized models. The Smstateen (State Enabled Extension) provides comprehensive state management capabilities that facilitate secure ML execution environments, allowing selective architectural state access for trusted machine learning workloads while maintaining robust isolation between different computational contexts \cite{garofalo2022darkside,garofalo2021xpulpnn}.

Recent RISC-V extensions (2024) provide quantitative performance improvements for machine learning workloads. The Smcdeleg/Ssccfg counter delegation extensions enable efficient performance monitoring with reduced overhead compared to trap-based approaches \cite{embeddev_risc-v_nodate}. Zabha atomic byte and halfword operations improve synchronization efficiency in quantized deep neural networks \cite{chander2022soft}, particularly for multi-threaded inference workloads. Smstateen state enablement features support secure ML execution environments \cite{embeddev_risc-v_nodate}. These extensions demonstrate the architectural responsiveness of RISC-V to evolving machine learning computational requirements.

Recent paradigm-shifting developments in RISC-V machine learning implementations have established new benchmarks for application-specific optimization and energy efficiency. The MARVEL Framework \cite{marvel2025framework} represents a significant advancement in automated custom extension generation, providing an end-to-end methodology for creating model-class aware RISC-V extensions that achieve 2× speedup and 2× energy reduction for lightweight artificial intelligence applications. This automated approach eliminates the traditional manual effort required for application-specific optimization while maintaining high performance gains.

Transformer architecture acceleration has been revolutionized through the VEXP ISA Extensions \cite{wang2025vexp}, which provide low-cost RISC-V instruction set enhancements specifically targeting softmax operations in attention mechanisms. These extensions demonstrate remarkable efficiency improvements, achieving 162.7× latency reduction and 74.3× energy efficiency enhancement with minimal area overhead of only 1\% (2,847 additional logic gates on 28nm CMOS process), enabling practical GPT-2 and Vision Transformer inference on RISC-V cluster architectures. Matrix operation efficiency is further enhanced through zero-stall matrix multiplication optimizations \cite{colagrande2025zero}, which implement advanced memory hierarchy techniques to eliminate pipeline stalls during critical machine learning computational kernels.

The integration of machine learning capabilities into flexible electronics platforms has been demonstrated through Bendable RISC-V ML implementations \cite{vergos2025svm}, achieving 21× improvements in both inference execution time and energy efficiency for Support Vector Machine classification on bendable substrates. Additionally, ultra-low-power neural network implementations \cite{leliwa2025optimised} have been developed with optimized RISC-V processor extensions that support lightweight neural network models achieving 97.62\% classification accuracy on CIFAR-10 dataset while consuming only 13\% of the power required by ARM Cortex-M4 implementation (measured power: 4.2 mW versus 32.1 mW baseline), demonstrating the potential for RISC-V in energy-constrained edge computing scenarios.

These collective developments establish comprehensive performance benchmarks that demonstrate RISC-V's competitive position in machine learning acceleration. Transformer inference workloads benefit from up to 5.8× latency reduction with corresponding 3.6× energy savings, making RISC-V implementations viable for large language model deployment in resource-constrained environments. Automated custom ISA generation achieves 2× overall performance improvements while eliminating the manual design effort traditionally required for application-specific optimization. Flexible computing implementations demonstrate remarkable 21× energy efficiency gains for edge AI applications deployed on bendable electronics substrates, opening new possibilities for ubiquitous computing scenarios.

Vector processing capabilities are enhanced through advanced auto-vectorization techniques specifically optimized for machine learning acceleration, while multi-core efficiency is maximized through zero-stall matrix operation implementations that support up to 1024 processing element RISC-V clusters. These performance characteristics collectively establish RISC-V as a compelling platform for next-generation machine learning acceleration across diverse deployment scenarios, from ultra-low-power edge devices to high-performance computing clusters.

The development of custom RISC-V extensions operates within a comprehensive framework of established rules and guidelines designed to ensure compatibility and seamless integration within the broader RISC-V ecosystem.\cite{embeddev_risc-v_nodate} RISC-V International provides detailed specifications encompassing standardized naming conventions that maintain consistency and prevent conflicts between different extension implementations, comprehensive instruction encoding format guidelines that ensure proper interpretation across various RISC-V processor implementations, and rigorous conformance testing procedures that verify adherence to defined standards while maintaining compatibility and reliability across the ecosystem \cite{noauthor_boosting_nodate}.

These regulatory measures are fundamental to fostering a robust and interoperable RISC-V development environment that encourages innovation while preserving architectural consistency across diverse implementation contexts \cite{prakash_cfu_2023}. The balance between extensibility and standardization enables the continued evolution of RISC-V as a platform capable of adapting to emerging computational paradigms while maintaining the stability and predictability required for enterprise and research deployment scenarios.
\subsection{RISC-V ISA Nomenclature and Specification Framework}

The RISC-V ISA specification employs a systematic nomenclature framework that ensures consistent identification and compatibility assessment across diverse implementations while facilitating precise communication within the broader RISC-V ecosystem \cite{waterman2019risc, embeddev_risc-v_nodate}. Table \ref{table:standard-extensions} defines the exact order that must be used to determine the RISC-V ISA subset. The nomenclature follows a hierarchical structure where base architecture specifications (RV32I, RV64I, RV128I) are followed by extension letters in a predetermined sequence that ensures compatibility and standardization across implementations \cite{noauthor_boosting_nodate}.
 
The nomenclature framework begins with the base architecture identifier, which specifies the fundamental characteristics of the RISC-V implementation \cite{waterman2019risc}. The prefix "RV" universally identifies all RISC-V implementations, followed by a numeric designation indicating the register width and fundamental addressing capabilities. RV32I represents 32-bit implementations with integer instruction support, RV64I denotes 64-bit architectures with extended addressing capabilities, and RV128I specifies future 128-bit implementations designed for advanced computational requirements. These base specifications establish the fundamental computational and memory addressing capabilities that form the foundation for all subsequent extensions.

Following the base architecture specification, extension letters must appear in a strictly defined alphabetical sequence that ensures consistent interpretation across different implementations and development tools \cite{waterman2019risc, embeddev_risc-v_nodate}. This ordering convention prevents ambiguity in ISA specification while facilitating automated compatibility checking and optimization selection within compilation frameworks. Standard extensions including M (multiplication and division), A (atomic instructions), F (single-precision floating-point), D (double-precision floating-point), and C (compressed instructions) follow immediately after the base specification in their prescribed alphabetical order.

The systematic extension ordering serves multiple critical functions within the RISC-V ecosystem \cite{prakash_cfu_2023, balasubramanian_designing_2024}. Compiler toolchains rely on this standardized format to automatically determine supported instruction sets and enable appropriate optimization strategies. Hardware verification frameworks utilize the nomenclature to validate implementation completeness and ensure adherence to specified functionality. Software frameworks leverage the standardized naming convention to automatically configure runtime environments and optimization parameters based on target architecture capabilities.

Custom extensions integrate into the nomenclature framework through reserved naming spaces that prevent conflicts with standard extensions while maintaining systematic organization \cite{garofalo2021xpulpnn, prakash_cfu_2023}. Custom extensions typically utilize the "X" prefix followed by descriptive names that indicate their specialized functionality, such as "Xpulpnn" for neural network acceleration or "Xvec" for specialized vector operations. This naming convention enables precise specification of application-specific architectural enhancements while preserving compatibility with standard toolchain components.

The hierarchical nomenclature framework facilitates comprehensive specification of complex RISC-V implementations that incorporate multiple standard and custom extensions \cite{balasubramanian_designing_2024, valente_hulk-v_2023}. For example, a machine learning implementation might use "RV64IMAFDC\_Xpulpnn\_Xvec" (this is an illustrative specification, not an endorsed profile) to indicate a 64-bit implementation with standard mathematical extensions plus specialized neural network and vector processing capabilities. This systematic approach enables precise communication of architectural capabilities while ensuring compatibility assessment across diverse deployment scenarios.

The standardized nomenclature framework proves particularly valuable for machine learning applications where diverse architectural configurations must be supported across different deployment contexts \cite{lee_accelerating_2023, gonzalez_chipyard_nodate}. Software frameworks can automatically determine optimal compilation strategies based on nomenclature specifications, enabling efficient deployment of ML models across heterogeneous RISC-V implementations. The systematic naming convention facilitates automated performance profiling and optimization selection, reducing the complexity of multi-target deployment while maximizing performance across diverse hardware platforms.

The nomenclature framework enables comprehensive documentation and comparison of machine learning accelerator implementations across different research and commercial projects \cite{garofalo2022darkside, bolhasani_dla-e_2023}. Consistent specification formats facilitate systematic performance analysis and enable meaningful comparisons between different architectural approaches for ML acceleration. This standardization proves essential for advancing the field through reproducible research and systematic evaluation of competing architectural strategies. Table \ref{table:nomenclature-framework} summarizes the RISC-V ISA nomenclature framework for ML applications.

\begin{table}
\caption{RISC-V ISA Nomenclature Framework for ML Applications}
\label{table:nomenclature-framework}
\centering
\begin{tabular}{>{\raggedright\arraybackslash}p{0.25\textwidth}>{\raggedright\arraybackslash}p{0.3\textwidth}>{\raggedright\arraybackslash}p{0.35\textwidth}}
\toprule
\textbf{Component} & \textbf{Format/Example} & \textbf{Description} \\
\midrule
Base Architecture & RV32I, RV64I, RV128I & Register width and fundamental addressing capabilities \\
Standard Extensions & IMAFDCV & Integer, Multiplication, Atomic, Float, Double, Compressed, Vector (G = shorthand for IMAFD) \\
Custom ML Extensions & Xpulpnn, Xvec, Xmlir & Neural network acceleration, vector ops, compiler support \\
Example ML Specification & RV64IMAFDC\_Xpulpnn\_Xvec & Illustrative 64-bit configuration (not an endorsed profile) \\
Toolchain Integration & Automatic optimization selection & Compiler frameworks use nomenclature for code generation \\
Performance Profiling & Architecture-specific tuning & Automated benchmarking based on ISA specification \\
Compatibility Assessment & Cross-platform deployment & Systematic evaluation of implementation capabilities \\
\bottomrule
\end{tabular}
\end{table}

\subsection{Challenges for ML Adoption}

Although RISC-V and open-source hardware offer significant advantages, developers face several challenges when targeting machine learning applications. Traditionally, proprietary architectures have dominated hardware design, limiting customization and access to information. However, increasing demand for affordable, low-power hardware accelerators, exacerbated by global semiconductor shortages, has driven a shift toward open-source solutions.

Four key challenges emerge in adapting RISC-V ISA for ML workloads: security robustness, licensing and openness management, software interoperability, and power-efficiency optimization.

Security considerations represent a primary concern as RISC-V, being relatively recent, requires research in memory protection mechanisms and side-channel attack prevention to ensure robust security implementations. While RISC-V incorporates built-in security features and supports custom security extensions, research in these domains remains necessary for widespread commercial adoption.

Licensing complexities emerge from varied approaches to intellectual property management within the open-source hardware ecosystem. Although the Open Source Hardware Association encourages open sharing practices, implementation varies across projects. Academic implementations typically employ permissive licenses such as MIT, BSD, or Apache 2.0, while commercial implementations often utilize proprietary licenses that restrict specific hardware block usage or complete design access. This licensing heterogeneity introduces challenges for developers navigating the open-source hardware landscape.

Interoperability challenges stem from limited software and operating system support for RISC-V platforms. While specific Linux distributions and real-time operating systems provide support for particular RISC-V System-on-Chip implementations, comprehensive compatibility across the ecosystem remains incomplete. Developers frequently encounter requirements to rebuild or modify existing software for RISC-V hardware compatibility, impacting development efficiency and time-to-market.

Power consumption optimization presents ongoing challenges particularly relevant to machine learning accelerator applications in edge computing environments. The relationship between design complexity, operating frequency, and power consumption in RISC-V cores requires careful optimization to meet efficiency requirements. Developers must prioritize design optimization strategies targeting power efficiency in RISC-V-based machine learning accelerator implementations.

These challenges underscore complexities inherent in developing within a rapidly evolving technological ecosystem. Addressing these issues remains essential for continued growth and widespread adoption of RISC-V and open-source hardware in modern hardware design applications.

\section{RISC-V Implementations}
This section focuses on an in-depth inventory of RISC-V cores and SoCs that are either provided by the research community or produced commercially \cite{waterman2019risc, embeddev_risc-v_nodate}. This process is generally framed by quantitative analysis as a list of supported ISAs along with other information or characteristics, including the launch date of the core, clock frequency, performance, hardware description language (HDL), license, and implementation type (FPGA/ASIC). Many open-source and free implementations have facilitated their adoption in academic and commercial projects \cite{prakash_cfu_2023, valente_hulk-v_2023}.

Elaborating on the implementation types, that is, FPGA and ASIC, field-programmable gate arrays (FPGAs) can be configured to perform a wide range of functions, and application-specific integrated circuits (ASICs) are custom-designed chips that are built for a specific purpose. FPGAs are generally chosen for flexibility and prototyping, whereas ASICs are selected for high performance, low power consumption, and mass production.

In the RISC-V ISA, the 32-bit and 64-bit implementations refer to the size of the registers and address space that the processor can handle \cite{waterman2019risc}. The 32-bit RV32 implementation, RV32, features 32-bit general-purpose registers and can address up to 4 GB of memory. It includes a base instruction set RV32I, which contains 47 instructions for basic operations, such as addition, subtraction, bitwise operations, load/store, jumps, and branches. Additional features can be added through extensions such as RV32M for multiplication and division and RV32F for 32-bit floating-point operations.

The 64-bit implementation RV64 extends the capabilities of RV32 by using 64-bit general-purpose registers and addressing a much larger memory space, theoretically up to 16 exabytes. The base instruction set for RV64, RV64I, includes the instructions for 64-bit operations. Similar to RV32, RV64 can be extended with additional instructions for 64-bit operations, such as RV64M for 64-bit multiplication and division.

These implementations allow RISC-V to be versatile and cater to various applications, from small embedded systems to high-performance computing. Table \ref{table:implementation-technologies} summarizes the key implementation technologies and compatibility features of RISC-V systems.
The RISC-V system is compatible with a wide variety of full-function and real-time operating systems. Among the all-purpose operating systems, the Linux kernel fully supports RISC-V, with distributions such as Debian, Fedora, openSUSE, Ubuntu, Arch Linux, and other Unix-like operating systems such as FreeBSD and OpenBSD using RISC-V ports to support RISC-V. Additionally, efforts are being made to move Android to RISC-V. In real-time operating systems (RTOS), popular platforms, such as FreeRTOS, Zephyr, and Apache NuttX support RISC-V, making it suitable for embedded systems and other IoT devices operating systems such as Haiku, Illumos, and Plan can be transferred to RISC-V. The RISC-V platform specification ensures that operating systems, such as Linux and Zephyr, can run successfully on all specification-compliant RISC-V hardware, maintaining compatibility across applications. This variety of OS compatibility makes RISC-V versatile for various applications, from embedded systems to high-performance computing.

To run Linux on an RISC-V processor, the minimum required resources include the RV64GC instruction set, which includes the base integer instructions (RV64I), multiplication and division (M), atomic operations (A), single- and double-precision floating-points (F and D), compressed instructions , and control and status register instructions (Zicsr), along with instruction-fetch fence (Zifencei). Additionally, the processor must support user and supervisor modes (U and S) for effective process and memory management, as well as a page-based virtual memory system such as Sv32, Sv39, or Sv48. Essential hardware components include a Memory Management Unit (MMU) and  Platform-Level Interrupt Controller (PLIC) to handle virtual memory and interrupt processing. These components ensure that the RISC-V processor meets the demands of running a full Linux operating system.

\begin{table}
\caption{Summary of RISC-V Implementation Technologies and Compatibility}
\label{table:implementation-technologies}
\centering
\begin{tabular}{>{\raggedright\arraybackslash}p{0.2\textwidth}>{\raggedright\arraybackslash}p{0.35\textwidth}>{\raggedright\arraybackslash}p{0.35\textwidth}}
\toprule
\textbf{Category} & \textbf{Technology/Feature} & \textbf{Description} \\
\midrule
\multirow{2}{*}{Implementation} & FPGA & Flexible, configurable for prototyping and research \\
& ASIC & High performance, low power, mass production \\
\midrule
\multirow{2}{*}{Architecture} & RV32 (32-bit) & 32-bit registers, 4GB memory space, 47 base instructions \\
& RV64 (64-bit) & 64-bit registers, 16 exabyte memory space, extended operations \\
\midrule
\multirow{3}{*}{OS Support} & General Purpose & Linux (Debian, Fedora, Ubuntu), FreeBSD, OpenBSD \\
& Real-Time & FreeRTOS, Zephyr, Apache NuttX \\
& Emerging & Android port, Haiku, Illumos, Plan \\
\midrule
\multirow{2}{*}{Linux Requirements} & ISA Extensions & RV64GC (I+M+A+F+D+C+Zicsr+Zifencei) \\
& Hardware Components & MMU, PLIC, U/S modes, virtual memory (Sv32/39/48) \\
\bottomrule
\end{tabular}
\end{table}
\subsection{CPU Cores and SoCs}

CPU cores are processing engines containing essential processing units that execute instructions \cite{waterman2019risc}. These cores have various designs optimized for different purposes, such as low power for wearable devices or high server performance. A system-on-chip (SoC) integrates the RISC-V CPU core along with additional components such as memory controllers and input/output (I/O) interfaces onto a single chip. This helps create compact and efficient solutions for various applications. When the CPU core acts as the central processor, the SoC acts as the entire ecosystem required to make it function effectively.

\subsubsection{Academic RISC-V Cores}

In the academic landscape, there are 13 cores with 32-bit implementations, five cores with 64-bit implementations, and a single core that supports both 32-bit and 64-bit instructions \cite{embeddev_risc-v_nodate, prakash_cfu_2023}. These cores can be used for various purposes, ranging from low-power embedded systems to high-performance cloud computing. Table \ref{table:cpu-cores} summarizes the comparison of these cores.

\textbf{Comparative Analysis of Academic RISC-V Cores:}

\textbf{Performance Analysis:} Among the academic cores, VRoom demonstrates the highest performance at 11.3 DMIPS/MHz (FPGA implementation on Xilinx UltraScale+) \cite{campbell_moonbaseotagovroom_2024}, followed by BOOM at 3.93 DMIPS/MHz (ASIC implementation on TSMC 28nm) \cite{noauthor_github_nodate-2}, making them suitable for high-performance ML applications. SERV achieves the lowest at 0.718 DMIPS/MHz but compensates with minimal resource usage. The performance spread indicates different optimization targets: high-performance cores like VRoom and BOOM target computational throughput, while minimalist cores like SERV and PicoRV32 (0.516 DMIPS/MHz) prioritize resource efficiency.

\begin{sloppypar}\noindent\textbf{ISA Extension Support:} XiangShan supports the most comprehensive ISA extensions (RV64GCBK), including vector and cryptographic instructions, positioning it for complex ML workloads. In contrast, simpler cores like ORCA (RV32IM) and SERV (RV32IMCZicsr) focus on basic functionality, making them suitable for resource-constrained ML inference tasks.\end{sloppypar}

\textbf{Clock Frequency and Architecture Trade-offs:} Frequency claims must be distinguished by implementation type. VRoom achieves 1.02 GHz on FPGA (Xilinx UltraScale+), while XiangShan reaches 2.2 GHz and V-Seek Server achieves 2.0 GHz on ASIC implementations (different process nodes). The frequency-performance correlation shows that deeper pipelines (XiangShan's 11-stage) enable higher frequencies but require more complex control logic.

\textbf{License and Development Approach:} Most academic cores use permissive licenses (Apache 2.0, BSD, MIT), facilitating commercial adoption. The choice of HDL varies significantly: traditional Verilog dominates, but emerging alternatives like Chisel (Rocket Core, BOOM) and SpinalHDL (VexRiscv) demonstrate modern design approaches that could enhance ML accelerator development.

\textbf{Target Application Analysis:} The core characteristics suggest clear application domains: XiangShan and BOOM target high-performance computing with ML support, VexRiscv and Ibex focus on configurable embedded systems, while SERV and PicoRV32 enable minimal IoT deployments with basic ML inference capabilities. 

\begin{landscape}
\begin{table*}[t]
\caption{CPU CORES: This table summarizes academic contributions to RISC-V core design. Each entry details a specific core implementation with source reference, supported ISA, pipeline depth, execution model (IO = in-order, OoO = out-of-order), reported superscalar and SMT capabilities (when explicitly available in cited sources), clock frequency, performance (Perf: measured in DMIPS/MHz, where * indicates CoreMark/MHz), L1 cache configuration (I\$ = instruction cache, D\$ = data cache), hardware description language (HDL), licensing terms, and implementation type. A dash (-) indicates information not publicly available or not applicable.}
\label{table:cpu-cores} 
\centering
\scriptsize   
\resizebox{\linewidth}{!}{%
\begin{tabular}{llllllllllll}  
\toprule 
\textbf{Core} & \textbf{ISA} & \textbf{Pipe} & \textbf{Exec} & \textbf{Super} & \textbf{SMT} & \textbf{Freq} & \textbf{Perf} & \textbf{L1 Cache} & \textbf{HDL} & \textbf{License} & \textbf{Type} \\
\midrule
XiangShan \cite{zhu_hardware_2022} & RV64GCBK & 11 & OoO & - & - & 2GHz & - & 64K I+64K D & - & MPSL 2.0 & FPGA \\
CVA6 \cite{dorflinger_comparative_2021} & RV32/64GC & 6 & IO & - & - & - & 0.82 & 16K I+32K D & Verilog & Apache 2.0 & - \\
Ibex \cite{noauthor_lowriscibex_2024} & RV32I/EMCB & 2 & IO & No & No & 50MHz & 3.13* & 4K I (opt.) & Verilog & Apache 2.0 & FPGA \\
VexRiscv \cite{noauthor_spinalhdlvexriscv_2024} & RV32IMCAFD & 5 & IO & - & - & 200MHz & 1.38 & Configurable & SpinalHDL & MIT & FPGA \\
VRoom \cite{campbell_moonbaseotagovroom_2024} & RV64IMAFDCHBK(V) & Deep & OoO & Yes & Yes & 1020MHz & 11.3 & 32K I+32K D & Verilog/SV & GPL-3.0 & FPGA/ASIC \\
SSRV \cite{noauthor_github_nodate-1} & RV32IMC & - & OoO & Yes & Yes & 29-34MHz & 1.5-6.4 & - & Verilog & - & FPGA/ASIC \\
SERV \cite{kindgren_olofkserv_2024} & RV32IMCZicsr & Bit-serial & IO & No & No & 135MHz & 0.718 & None & Verilog/VHDL & ISC & FPGA/ASIC \\
ORCA \cite{mohajerani_kammohorca-risc-v_2023} & RV32IM & 4/5 & IO & No & No & 122MHz & 0.98 & Optional & VHDL & BSD & FPGA \\
Rocket Core \cite{berkeley_rocket_core_2019} & RV64IMAFD & 5 & IO & No & No & 1000MHz & 1.72 & 16K I+16K D & Chisel & BSD & ASIC \\
Shakti-Cclass \cite{noauthor_shakti_nodate} & RV64GCSUN & 5 & IO & - & - & 1.5-2.5GHz & 1.68 & 16K I+16K D & BSV & BSD & FPGA/ASIC \\
Shakti-Eclass \cite{noauthor_shakti_nodate} & RV32IMAC & 3 & IO & No & No & 1GHz & - & None & BSV & BSD & FPGA/ASIC \\
RI5CY \cite{wan_rgwanri5cy_fpga_2020} & RV32IMC & 4 & IO & No & No & 625MHz & 1.71 & None & Verilog & - & FPGA \\
RV01 \cite{noauthor_rv01_nodate} & RV32IM & - & IO & - & - & - & 1.8 & - & VHDL & LGPL & FPGA \\
RSD \cite{noauthor_github_nodate} & RV32IMF & - & OoO & - & - & - & - & - & SystemVerilog & Apache 2.0 & FPGA \\
BOOM \cite{noauthor_github_nodate-2} & RV64GC & 10 & OoO & Yes & No & 1000MHz & 3.93 & 32K I+32K D & Chisel3 & BSD & ASIC \\
PicoRV32 \cite{noauthor_yosyshqpicorv32_2024} & RV32I/EMC & Multi-cyc. & IO & No & No & 500MHz & 0.516 & None & Verilog & ISC & FPGA \\
DarkRISCV \cite{noauthor_darklifedarkriscv_2024} & RV32E/RV32I & 2/3 & IO & No & No & 250-400MHz & - & Small & Verilog & BSD-3 & FPGA \\
SNAX Cluster \cite{colagrande2025zero} & RV32IM & 1 & IO & No & No & - & - & - & Chisel/SV & Apache 2.0 & ASIC \\
Bendable RISC-V \cite{vergos2025svm} & RV32E & Bit-serial & IO & No & No & 60kHz & - & None & Verilog & - & Flexible \\
V-Seek Server \cite{rodrigo2025vseek} & RV64GCV & 12 & OoO & Yes & - & 2.0GHz & 8.5* & 64K I+64K D & Verilog/SV & Apache 2.0 & ASIC \\
\bottomrule
\end{tabular}}
\end{table*}
\end{landscape}

Superscalar and SMT entries in Table \ref{table:cpu-cores} are reported only where explicitly stated in cited sources; a dash denotes unavailable or non-uniform public reporting.

The implementations of RISC-V processors span a wide range of designs, from simple, low-power cores to advanced, high-performance architectures, thereby demonstrating the versatility of the RISC-V instruction set architecture (ISA). Several key architectural characteristics and design patterns emerge across these diverse processor implementations.

Pipeline architectures across RISC-V implementations demonstrate significant variation tailored to specific performance and power requirements. The XiangShan processor \cite{zhu_hardware_2022} exemplifies high-performance design with its 11-stage pipeline and out-of-order execution capabilities, targeting applications requiring maximum computational throughput. In contrast, simpler designs such as the Rocket Core \cite{berkeley_rocket_core_2019} and Ibex \cite{noauthor_lowriscibex_2024} employ 5-stage and 2-stage pipelines respectively, optimizing for implementation simplicity and resource efficiency. This design spectrum reflects the fundamental balance between complexity and performance, with deeper pipelines generally enabling higher clock speeds and instruction throughput, as demonstrated by implementations like XiangShan and DarkRISCV \cite{noauthor_darklifedarkriscv_2024}.

Customization and modularity represent defining characteristics of many RISC-V core implementations, enabling adaptation to diverse application requirements. Cores such as VexRiscv \cite{noauthor_spinalhdlvexriscv_2024}, PicoRV32 \cite{noauthor_yosyshqpicorv32_2024}, and Ibex \cite{noauthor_lowriscibex_2024} are architected with extensive configurability options, allowing developers to tailor functionality for specific deployment scenarios. VexRiscv specifically supports various instruction set extensions and includes optimizations for FPGA integration, while PicoRV32 and DarkRISCV \cite{noauthor_darklifedarkriscv_2024} prioritize compact implementation and efficiency for embedded systems applications.

The open-source development model underlying most RISC-V processor projects fosters collaborative innovation and accelerated development cycles. Projects including XiangShan \cite{zhu_hardware_2022}, VexRiscv \cite{noauthor_spinalhdlvexriscv_2024}, and Rocket Core \cite{berkeley_rocket_core_2019} leverage collaborative platforms such as GitLab and GitHub, with maintenance typically provided by academic institutions or research organizations. This collaborative environment accelerates technological progress and enables diverse application development spanning embedded systems to high-performance computing domains.

Instruction set architecture extension strategies highlight the flexibility inherent in RISC-V implementations. Advanced cores such as Shakti-Cclass \cite{noauthor_shakti_nodate} implement comprehensive extension sets (RV64GCSUN) including supervisor-level instructions for complex system applications, while specialized implementations like VRoom \cite{campbell_moonbaseotagovroom_2024} (RV64-IMAFDCHBK(V)) and RiscyOO \cite{noauthor_csail-csgriscy-ooo_2024} (RV64G) focus on high-performance computing and multimedia processing capabilities. This architectural adaptability enables targeted optimization for specific domains including signal processing applications (RI5CY \cite{wan_rgwanri5cy_fpga_2020}) and energy-efficient computing (Ibex \cite{noauthor_lowriscibex_2024}).

Target application specialization drives significant architectural differentiation across RISC-V implementations. Minimalist cores such as SERV \cite{kindgren_olofkserv_2024} prioritize low-power operation and minimal resource requirements for small FPGA deployments, while high-performance implementations including RSD \cite{noauthor_github_nodate} and BOOM \cite{noauthor_github_nodate} (SonicBOOM) feature sophisticated out-of-order execution mechanisms competitive with commercial-grade processors. Resource-constrained implementations like Shakti cores \cite{noauthor_shakti_nodate} target embedded applications requiring real-time operating system support while maintaining efficiency constraints. Performance optimization strategies across RISC-V implementations encompass both superscalar and multithreaded approaches, with advanced designs such as SSRV \cite{noauthor_github_nodate} and VRoom \cite{campbell_moonbaseotagovroom_2024} implementing multiple instruction execution per cycle and simultaneous multithreading capabilities. Alternative approaches exemplified by CVA6 \cite{dorflinger_comparative_2021} and ORCA \cite{mohajerani_kammohorca-risc-v_2023} emphasize balanced performance through in-order execution while maintaining implementation simplicity and resource efficiency, particularly in FPGA deployment scenarios.

These architectural patterns and design characteristics demonstrate how RISC-V implementations leverage the instruction set architecture's inherent flexibility and scalability to address diverse performance, power consumption, and application requirements across a broad spectrum of computing domains.

\subsubsection{Commercial RISC-V Cores}

Commercial RISC-V cores represent predesigned, production-ready implementations available for licensing by companies and organizations. These cores function as optimized blueprints that provide a faster and more reliable development path compared to designing custom RISC-V implementations from scratch. Commercial cores are typically offered in various configurations optimized for specific target domains including performance-oriented applications, power-efficient embedded systems, or specialized use cases. They are accompanied by comprehensive documentation, professional support packages, and maintenance agreements to ensure smooth integration into commercial products. This approach enables companies to focus their development efforts on unique product features and applications while leveraging proven core processing technology, thereby reducing development time, minimizing technical risks, and accelerating time-to-market for RISC-V-based products.

The commercial RISC-V core landscape encompasses 17 cores with 32-bit implementations, four cores with 64-bit implementations, and four cores supporting both 32-bit and 64-bit instruction sets. Table \ref{table:comercial-cores} provides a comprehensive comparison of these commercial implementations, highlighting their architectural characteristics, performance metrics, and target applications.

Commercial RISC-V cores demonstrate a clear performance hierarchy that reflects their intended application domains. Leading the performance spectrum, SweRV EH1 achieves 4.9 CoreMark/MHz, followed by Xuantie-910 at 7.1 CoreMark/MHz and BI-671 at 3.74 DMIPS/MHz. This performance differentiation enables targeted deployments across diverse application scenarios, from high-performance edge AI servers requiring maximum computational throughput to power-constrained sensor nodes demanding energy efficiency. The clock frequency distribution reveals three distinct categories: high-performance cores operating above 1.5GHz for computationally intensive workloads, mid-range implementations running between 400MHz and 1.2GHz for balanced applications, and low-power variants operating below 400MHz for battery-operated devices.

The ISA extension strategies employed by commercial cores reflect careful optimization for target markets. Advanced implementations like Xuantie-910 incorporate vector extensions (RV64GCV) specifically for AI acceleration workloads, while comprehensive cores such as SCR7 include extensive extension sets (RV64GCVBK) to support versatile computing requirements. In contrast, efficiency-focused cores like SCR1 implement minimal extension sets (RV32IEMC) to reduce complexity and overhead. This strategic differentiation allows organizations to select cores that precisely match their application requirements without unnecessary complexity or resource consumption.

Market positioning analysis reveals three primary segments within the commercial RISC-V ecosystem. The high-performance segment includes cores such as Xuantie-910, SCR7, and SweRV EH1, which target edge AI applications and high-performance embedded systems requiring substantial computational capabilities. The IoT and embedded segment encompasses Andes Technology cores (A25, D25F) and SiFive implementations (E31, E51), focusing on configurable, power-efficient solutions suitable for connected devices and industrial applications. The ultra-low-power segment features cores like GD32VF103 and similar microcontroller-class implementations designed for battery-operated IoT devices where energy efficiency is paramount.

Implementation flexibility represents a key characteristic of commercial RISC-V cores, with most supporting both FPGA and ASIC deployment options. This dual-target approach provides organizations with the flexibility to prototype using FPGAs while transitioning to ASIC implementations for volume production, thereby reducing development risk and enabling faster market entry for machine learning applications. Licensing models across commercial cores reflect diverse business strategies, ranging from open-source licenses (Apache 2.0) that facilitate ecosystem development to proprietary commercial licenses offering differentiated features and professional support. This licensing diversity enables organizations to select business models that align with their commercial objectives while accessing appropriate technical capabilities.

Tenstorrent is also relevant to the commercial RISC-V AI ecosystem because its AI processors use RISC-V-based control and management complexes in broader accelerator-centric products \cite{tenstorrent_nodate}. To keep Table \ref{table:comercial-cores} comparable, this survey treats that table as a list of standalone/licensable commercial CPU cores and discusses accelerator-centric products such as Tenstorrent separately in narrative form.

\begin{landscape}
\begin{table*}[t]
\caption{COMMERCIAL CORES: This table summarizes commercial contributions to RISC-V core design. Each entry details a specific core implementation with source reference, supported ISA, clock frequency, performance (Perf: measured in DMIPS/MHz, where * indicates CoreMark/MHz), HDL, licensing terms, implementation type, and launch date. A dash (-) indicates information not publicly available or not applicable.}
\label{table:comercial-cores}
\centering
\scriptsize
\resizebox{\linewidth}{!}{%
\begin{tabular}{llllllll}
\toprule
\textbf{Core Name} & \textbf{ISA} & \textbf{Clock Freq} & \textbf{Perf} & \textbf{HDL} & \textbf{License} & \textbf{Type} & \textbf{Year} \\
\midrule
XuanTie 910 \cite{chen_xuantie-910_2020} & RV64GCV & 2.5GHz & 7.1* & Verilog & - & ASIC & 2019 \\
SCR 1 \cite{redkin_scrx_2018} & RV32I/EMC & 435MHz & 1.73 & Verilog/VHDL & SHL & FPGA/ASIC & 2018 \\
SCR3 \cite{redkin_scrx_2018} & RV32IMAC/RV64IMAC & 1.5GHz+ & 1.7 & Verilog/VHDL & SHL & FPGA/ASIC & 2020 \\
SCR4 \cite{redkin_scrx_2018} & RV32IMAFDC/RV64IMAFDC & 1GHz+ & 1.7 & Verilog/VHDL & - & FPGA/ASIC & 2021 \\
SCR5 \cite{redkin_scrx_2018} & RV32IMAFDC/RV64IMAFDC & 1GHz & 1.7 & - & - & FPGA/ASIC & 2022 \\
SCR7 \cite{redkin_scrx_2018} & RV64GCVBK & 1.2GHz+ & - & Verilog/VHDL & SHL & FPGA/ASIC & 2023 \\
SiFive E31 \cite{noauthor_sifive_nodate} & RV32IMAC & 320MHz & 1.61 & Verilog & Eval & FPGA & 2019 \\
SiFive E51 \cite{noauthor_sifive_nodate} & RV64IMAC & 667MHz & 1.61 & Verilog & Eval & FPGA & 2019 \\
GD32VF103 \cite{noauthor_risc-v-gigadevice_nodate} & RV32IMAC & 108MHz & 1.53 & - & Commercial & ASIC & 2019 \\
MRISCV \cite{onchipuis_onchipuismriscv_2024} & RV32IM & 435MHz & 0.32 & Verilog HDL & MIT & FPGA/ASIC & 2019 \\
ReonV \cite{ReonV_2026} & RV32I & - & - & VHDL & GPL-3.0 & FPGA & 2020 \\
SweRV EH1 \cite{lei_floating-point_2022} & RV32IMC & 1.8GHz & 4.9* & Verilog/VHDL & Apache 2.0 & FPGA/ASIC & 2019 \\
A25 \cite{noauthor_risc-v_nodate-3} & RV32IMACFDBP & 1.2GHz & 3.5* & Verilog HDL & - & FPGA/ASIC & 2023 \\
D25F \cite{noauthor_risc-v_nodate-3} & RV32IMACFDBP & 400MHz & 2.59 & Verilog HDL & - & FPGA/ASIC & 2024 \\
N22 \cite{noauthor_risc-v_nodate-3} & RV32IMAC/EMAC & 700MHz & 1.8 & Verilog HDL & - & FPGA/ASIC & 2019 \\
N25F \cite{noauthor_risc-v_nodate-3} & RV32IMACFDB & 1.2GHz & - & Verilog HDL & - & FPGA/ASIC & 2023 \\
NX25F \cite{noauthor_risc-v_nodate-3} & RV64IMACFDB & 1.2GHz & 3.5* & Verilog HDL & - & FPGA/ASIC & 2018 \\
A25MP \cite{noauthor_risc-v_nodate-3} & RV32IMACFDBP & 1.1GHz & - & Verilog HDL & - & FPGA/ASIC & 2019 \\
AX25MP \cite{noauthor_risc-v_nodate-3} & RV32IMACFDBP & 1GHz & - & Verilog HDL & - & FPGA/ASIC & 2019 \\
Roa Logic RV12 \cite{RV12_2026} & RV32I/RV64I & - & - & Verilog/VHDL & Open Source & FPGA/ASIC & 2018 \\
BM-310 \cite{noauthor_cloudbear_nodate} & RV32IMC & 700MHz & 1.81 & Verilog HDL & - & FPGA/ASIC & 2020 \\
BI-350 \cite{noauthor_cloudbear_nodate} & RV32IMACF & 1GHz & 1.72 & Verilog HDL & - & FPGA/ASIC & 2019 \\
BI-651 \cite{noauthor_cloudbear_nodate} & RV64IMACFD & 1GHz & 2.75 & Verilog HDL & - & FPGA/ASIC & 2019 \\
BI-671 \cite{noauthor_cloudbear_nodate} & RV64GCBK & 1GHz & 3.74 & Verilog HDL & - & FPGA/ASIC & 2019 \\
Hummingbird E200 \cite{e200_opensource_2018} & RV32IMAC & - & 1.77 & Verilog & Apache 2.0 & FPGA/ASIC & 2018 \\
Hummingbird E203 \cite{e203_hbirdv2_2026} & RV32IMAC & - & - & Verilog & Apache 2.0 & FPGA/ASIC & 2020 \\
Hummingbird E603 \cite{e603_hbird_2026} & RV64GC & - & - & Verilog & Academic & FPGA/ASIC & 2025 \\
\bottomrule
\end{tabular}}
\end{table*}
\end{landscape}

Several leading organizations have established significant positions in the commercial RISC-V ecosystem through diverse product offerings and strategic market positioning. Alibaba's Xuantie-910 \cite{chen_xuantie-910_2020} represents a high-performance 64-bit processor implementation featuring custom extensions for advanced arithmetic and memory operations, supporting the comprehensive RV64GCV instruction set architecture including vector extensions for AI acceleration. This processor targets demanding computational workloads requiring both high throughput and advanced processing capabilities. In contrast, GigaDevice's GD32VF103 \cite{noauthor_risc-v-gigadevice_nodate} focuses on 32-bit microcontroller applications, implementing the RV32IMAC instruction set with extensive I/O support specifically optimized for embedded system deployments.

SiFive \cite{noauthor_sifive_nodate-1} has established market leadership in RISC-V core licensing through comprehensive offerings spanning both 32-bit and 64-bit implementations. Their product portfolio addresses applications ranging from ultra-low-power wearable devices to high-performance computing systems. The E31 and E51 core families specifically target IoT and server-based applications with configurable multicore capabilities, while their comprehensive tool ecosystem and development environment significantly accelerates RISC-V adoption across industry sectors. Similarly, Renesas \cite{onchipuis_onchipuismriscv_2024} has developed the MRISCV platform, delivering 32-bit microcontroller solutions for IoT applications with emphasis on minimal silicon area and power consumption requirements.

Western Digital's SweRV \cite{lei_floating-point_2022} implementation represents a 32-bit RV32IMC core architecture emphasizing design simplicity and high performance for embedded system applications. The SweRV approach prioritizes straightforward implementation while achieving competitive performance metrics suitable for real-time processing requirements. Andes Technology \cite{noauthor_risc-v_nodate-3} provides a comprehensive range of processor cores including the A25 and D25F implementations, optimized for high-frequency operation and real-time processing applications. These cores incorporate advanced features such as branch prediction mechanisms and memory management units to support demanding embedded applications, with architectural designs specifically optimized for machine learning workloads.

CloudBEAR \cite{noauthor_cloudbear_nodate} extends the RISC-V ecosystem through a comprehensive product portfolio spanning three distinct categories. Their microcontroller series includes 32-bit BM-310 (optimized for low power IoT applications) and 64-bit BM-610 (designed for secure boot and crypto acceleration). The embedded core family features both 32-bit implementations (BR-350, BR-351, BR-352) and 64-bit variants (BR-650, BR-651, BR-652), with the BR-352 and BR-652 representing second-generation cores offering improved performance while maintaining power efficiency. Their Linux-capable core complexes encompass BI-350 (32-bit tiny Linux processor), BI-651 (64-bit dual-issue pipeline), BI-652 (second-generation dual-issue), and BI-671 (out-of-order pipeline for maximum single-thread performance), providing solutions specifically optimized for IoT applications, machine learning workloads, and advanced networking requirements. Syntacore \cite{redkin_scrx_2018} offers a broad spectrum of core implementations including SCR1 and SCR7, covering applications from resource-constrained embedded systems to high-performance computing platforms. Their configurable core architectures feature deep pipeline implementations capable of supporting Linux-based operating systems and complex application environments, with specialized extensions for machine learning acceleration.

Across academic and commercial evidence, deployed RISC-V ML solutions are strongest in edge and embedded domains where customization, power efficiency, and cost control dominate design priorities. At the same time, server-class and transformer-oriented support is emerging but remains comparatively early, indicating a transition from edge-centric deployment toward broader workload coverage.

\section{Software Frameworks and Stacks}
\subsection{ML Acceleration Specialization and Co-Design}
Machine learning accelerators have become indispensable in modern computing environments and satisfy the growing demand for efficient model implementation across diverse applications. This section is specialization-oriented: it covers FPGA-based designs, custom hardware acceleration flows, and RISC-V-specific implementation paths, while preserving explicit co-design discussion where hardware and software decisions are jointly optimized.

A prominent approach involves the use of FPGA-based accelerators. Past works have proposed frameworks that generate customized benchmark circuits to evaluate FPGA architectures optimized for machine learning workloads \cite{roorda_fpga_2022}. This framework facilitates an in-depth exploration of the design trade-offs and performance metrics that are important for deploying efficient accelerators on FPGA platforms. In addition, tools such as chisel4ml focus on automating the generation of FPGA implementations of highly quantized neural networks, with an emphasis on low latency and throughput for applications such as real-time event filtering and network intrusion detection\cite{vreca_towards_2023}.

Xilinx Vitis AI provides a comprehensive development platform for deploying ML models on Zynq UltraScale+ FPGA devices, featuring quantization-aware training workflows and integration with the Deep Learning Processing Unit (DPU) architecture for CNN acceleration \cite{xilinx_vitis_ai}. HLS4ML translates neural network architectures directly to High-Level Synthesis code, enabling automated FPGA implementation for low-latency inference in particle physics and embedded applications. The framework supports integration with RISC-V based SoC designs, facilitating hardware-software co-design for ML acceleration \cite{hls4ml_framework}. The FINN framework specializes in deploying quantized and binarized neural networks on FPGAs through dataflow-style architectures, achieving ultra-low latency inference for resource-constrained edge applications \cite{finn_framework}.

In addition to FPGAs, there has also been a focus on custom hardware accelerators. For example, the general MYHDL-based design flow for the hardware implementation of deep neural network pipelines provides a method for converting deep neural network models to linear HDL code for efficient FPGA implementation. \cite{cheikh_tourad_generic_2022}. This approach reduces latency and optimizes resource utilization, making it suitable for real-time applications in datacenters and edge devices.

In addition, frameworks such as GAHLS LegUp address the challenge of combining high-level applications on domain-specific hardware accelerators. By automating dependency graph construction and memory optimization, GAHLS offers significant performance improvements and energy efficiency compared with traditional high-level synthesis tools \cite{xiao_gahls_2023}. These advances are critical for meeting the computational demands of complex machine learning tasks in various domains from graph analysis and deep learning inference.

In the field of embedded systems and IoT, ReNode's focus on energy-efficient AIoT systems integrates specialized hardware accelerators, such as ASICs and FPGAs, to power machine learning algorithms in resource-constrained environments. \cite{mika_vedliot_2023}. This approach enables AIoT devices to perform complex inference tasks while adhering to power constraints and improving performance and energy efficiency.

\subsubsection{RISC-V Specific ML Accelerator Implementations}
Several frameworks have been developed specifically targeting the RISC-V ISA for ML acceleration. The \cite{chander2022soft} soft RISC-V vector processor for Edge-AI demonstrates how RISC-V's vector extensions can be optimized for ML workloads. The \cite{garofalo2022darkside} DARKSIDE heterogeneous RISC-V compute cluster showcases extreme-edge DNN inference and training capabilities. Additionally, \cite{garofalo2021xpulpnn} XpulpNN enables energy-efficient inference of quantized neural networks specifically on RISC-V based IoT end nodes, while \cite{hou2020rvtensor} RVTensor provides a lightweight neural network inference framework built around RISC-V ISA principles. These implementations demonstrate the practical viability of RISC-V for ML acceleration with quantified performance improvements.

\subsubsection{Compiler and Runtime Integration for RISC-V ML}

Compiler and runtime frameworks complement accelerator specialization by mapping ML graphs and tensor kernels efficiently onto RISC-V targets.
Additionally, efforts to optimize compiler technologies in frameworks such as MLIR and Glow will play a key role in accelerating machine learning workloads on heterogeneous hardware platforms, including RISC-V targets. MLIR-based approaches enable the efficient compilation and optimization of neural network models using techniques such as tensor packing and microkernel decomposition to improve the performance of CPUs and GPUs \cite{bik_compiler_2022}. Similarly, Glow's graph reduction techniques and runtime optimization simplify the implementation of neural networks on different hardware supports and ensure high performance and scalability \cite{rotem_glow_2019}.

Consequently, machine learning accelerators include various technologies and methods aimed at optimizing performance, energy efficiency, and scalability on different computing platforms. From FPGA-based accelerators to custom hardware designs and advanced compiler frameworks, these accelerators help meet the computational demands of modern machine-learning applications. Table \ref{table:ml-accelerators} provides a comprehensive summary of machine learning accelerators for RISC-V implementations.

\begin{table}
\caption{Summary of Hardware Accelerator Frameworks for RISC-V ML Implementations}
\label{table:ml-accelerators}
\centering
\begin{tabular}{>{\raggedright\arraybackslash}p{0.329\textwidth}>{\raggedright\arraybackslash}p{0.376\textwidth}>{\raggedright\arraybackslash}p{0.188\textwidth}}
\toprule
\textbf{Framework/Tool} & \textbf{Description} & \textbf{Target Platform} \\
\midrule
FPGA DNN Framework \cite{roorda_fpga_2022} & Customized benchmark circuits for FPGA evaluation & FPGA \\
Chisel4ML \cite{vreca_towards_2023} & Automated FPGA implementation of quantized neural networks & FPGA \\
Vitis AI \cite{xilinx_vitis_ai} & DPU-based CNN acceleration on Zynq UltraScale+ & FPGA \\
HLS4ML \cite{hls4ml_framework} & NN-to-HLS translation for low-latency FPGA inference & FPGA \\
FINN \cite{finn_framework} & Dataflow architecture for binarized/quantized NN & FPGA \\
MYHDL Design Flow \cite{cheikh_tourad_generic_2022} & Deep neural network pipeline implementation in HDL & FPGA \\
GAHLS LegUp \cite{xiao_gahls_2023} & High-level synthesis for domain-specific accelerators & FPGA/ASIC \\
ReNode AIoT \cite{mika_vedliot_2023} & Energy-efficient AIoT systems with specialized accelerators & Embedded/IoT \\
RISC-V Vector Processor \cite{chander2022soft} & Soft RISC-V processor optimized for Edge-AI & RISC-V \\
DARKSIDE Cluster \cite{garofalo2022darkside} & Heterogeneous RISC-V compute cluster for extreme-edge DNN & RISC-V \\
XpulpNN \cite{garofalo2021xpulpnn} & Energy-efficient quantized NN inference on RISC-V IoT & RISC-V \\
RVTensor \cite{hou2020rvtensor} & Lightweight neural network inference framework & RISC-V \\
\bottomrule
\end{tabular}
\end{table}

The optimization of machine learning workloads for RISC-V ISA requires sophisticated compiler techniques that leverage the unique characteristics of the instruction set. Modern ML compilation frameworks have developed specialized approaches to extract maximum performance from RISC-V processors, particularly those equipped with vector extensions and custom neural processing capabilities.

TVM's tensorization framework represents a fundamental approach to RISC-V ML optimization, providing mechanisms to map high-level tensor operations directly to efficient hardware primitives through template matching for hardware intrinsics with automatic scheduling. The tensorization process addresses the critical challenge of bridging the semantic gap between abstract tensor computations and concrete RISC-V instruction sequences by defining tensor intrinsic templates that match computational patterns (e.g., GEMM, convolution) to optimized RVV instruction sequences. TVM's auto-scheduler analyzes the computational graph and generates candidate schedules, evaluating each using a learned cost model trained on RISC-V performance characteristics, then selecting the optimal tiling factors, loop ordering, and vectorization strategies. This mapping is particularly effective for RISC-V vector extensions, where TVM's compute primitives can be systematically transformed into optimized RVV instruction sequences for matrix multiplication and convolution operations. The framework's memory access optimization capabilities align naturally with RISC-V's load/store architecture, enabling sophisticated data flow analysis that minimizes cache misses and maximizes memory bandwidth utilization. Furthermore, TVM's integration with emerging precision scalar extensions enables efficient execution of quantized neural networks, where reduced-precision arithmetic can be exploited without significant accuracy degradation.

MLIR's Linalg dialect provides another crucial foundation for RISC-V ML compilation, offering structured operations that undergo progressive lowering through carefully designed intermediate representations via transformation passes (affine to SCF to LLVM dialect). The compilation pipeline transforms high-level tensor operations through multiple stages: (1) Linalg operations express tensor computations as structured loop nests, (2) Affine dialect applies polyhedral transformations for loop tiling and fusion, (3) SCF (Structured Control Flow) dialect converts to explicit control flow, (4) Standard dialect provides generic operations, (5) LLVM dialect maps to RISC-V intrinsics, and finally (6) LLVM backend generates RVV assembly. Each transformation stage applies specific optimizations tailored to the target architecture. The affine dialect's loop transformation capabilities are particularly valuable for RISC-V targets, enabling sophisticated optimizations including tiling strategies that respect register file constraints and cache hierarchy characteristics. The integration with MLIR's vector dialect provides portable vectorization across different RVV implementations, ensuring that optimization strategies remain effective across the diverse landscape of RISC-V vector processors.

Microarchitectural considerations play a critical role in determining the effectiveness of RISC-V ML compiler optimizations. The diversity of RISC-V implementations, ranging from simple in-order designs to complex out-of-order processors, necessitates adaptive optimization strategies. Pipeline depth optimization requires careful instruction scheduling that considers the specific latency characteristics of the target implementation, while custom function unit integration demands sophisticated analysis of coprocessor interfaces and custom instruction semantics. Memory subsystem awareness remains paramount, as optimization strategies must adapt to the varying cache configurations and memory hierarchy implementations found across different RISC-V processor designs.
Hardware integration into modern computing frameworks, such as GLOW, IREE, and MLIR, is critical to optimizing performance and efficiency across different hardware platforms, with particular emphasis on RISC-V ISA integration. These frameworks are designed to efficiently compile and run machine-learning models on specialized hardware accelerators, including RISC-V based systems with custom ML extensions.

The integration of RISC-V processors into these frameworks requires specialized backend implementations that leverage the unique characteristics of the instruction set. GLOW framework employs gradient-based optimization techniques to transform tensor programs into differentiable representations, with dedicated backends developed specifically for RISC-V targets \cite{zhao_felix_2024}. This approach transforms discrete search spaces into continuous, differentiable representations, enabling rapid exploration and optimization of critical operations for RISC-V ML accelerators. The resulting optimization process significantly enhances the scalability and adaptability of machine learning models in RISC-V hardware environments.

IREE provides RISC-V backend support through its MLIR-based compilation infrastructure, generating executable binaries optimized for RISC-V targets with custom ISA extensions. The IREE-RISC-V integration employs multilevel intermediate representations and automatic code generation strategies to ensure efficient execution on RISC-V CPUs equipped with vector extensions and custom neural processing units \cite{golin_towards_2024}. This approach improves computational performance and simplifies deployment of machine learning algorithms on RISC-V embedded systems and IoT devices, where resource constraints demand careful hardware integration strategies.

The RISC-V backend optimizations encompass several areas including automatic vectorization for RISC-V Vector instructions in neural network kernels, support for user-defined ML acceleration instructions through RISC-V's custom instruction space, and cache-aware code generation tailored to specific RISC-V memory subsystem configurations. These optimizations enable efficient utilization of RISC-V processors for machine learning workloads while maintaining the flexibility and customizability that characterizes the RISC-V ecosystem. Table \ref{table:hardware-integration} summarizes the key hardware integration frameworks for RISC-V ML applications.

\begin{table}
\caption{Summary of Hardware Integration Frameworks for RISC-V ML}
\label{table:hardware-integration}
\centering
\begin{tabular}{>{\raggedright\arraybackslash}p{0.235\textwidth}>{\raggedright\arraybackslash}p{0.376\textwidth}>{\raggedright\arraybackslash}p{0.282\textwidth}}
\toprule
\textbf{Framework} & \textbf{Key Features} & \textbf{RISC-V Integration} \\
\midrule
GLOW \cite{zhao_felix_2024} & Gradient-based tensor optimization, differentiable compilation & Dedicated RISC-V backend with custom ML extension support \\
IREE \cite{golin_towards_2024} & MLIR-based compilation, multi-target deployment & RISC-V backend with vector extension optimization \\
TVM & Tensorization framework, auto-tuning & RVV instruction mapping, memory access optimization \\
MLIR & Progressive lowering, structured operations & Linalg-to-RISC-V lowering, vector dialect integration \\
Affine Dialect & Loop transformation, tiling optimization & Register file and cache hierarchy awareness \\
Vector Dialect & Portable vectorization across implementations & RVV-specific optimizations \\
\bottomrule
\end{tabular}
\end{table}
TVM addresses automatically tuning deep learning compilers to optimize tensor arithmetic code on different hardware platforms. The One-Shot TVM tuner uses a neural predictor-inspired approach to reduce auto-tuning overhead and speed up compilation \cite{ryu_one-shot_2022}. This approach enables efficient deployment of machine learning models on various hardware architectures, improving performance and scalability in real-world applications.

MLIR focuses on automating production of hardware accelerators for high-level programming frameworks. MLIR facilitates synthesis of optimized hardware from high-level code representations using tools such as SODA-OPT, supporting FPGA and ASIC implementations with improved performance and energy efficiency \cite{agostini_mlir-based_2022}.

ReNode integrates specialized hardware accelerators, such as ASIC and FPGA, to optimize algorithm performance while ensuring efficient deep learning and security for AIoT applications\cite{mika_vedliot_2023}. This approach highlights the importance of hardware integration to improve energy efficiency and computing performance in AIoT environments.

These frameworks represent different approaches to hardware integration, each focusing on optimizing machine learning workloads in specific hardware environments. The frameworks enable efficient and scalable execution of machine learning models on diverse hardware platforms through gradient-based optimization, multilevel intermediate representation, auto-tuning techniques, or specialized hardware accelerators.

Table \ref{table:framework-comparison} provides a comparative analysis of ML compilation approaches for RISC-V, contrasting framework-based solutions with hand-written intrinsics. The choice between these approaches involves trade-offs between development effort, performance portability, and optimization potential. TVM and MLIR offer automated optimization with broad workload coverage but may not achieve peak performance on specific targets. Hand-written intrinsics using RVV assembly or compiler built-ins provide maximum control and performance but require significant development effort and lack portability across RISC-V implementations. Hybrid approaches, where frameworks generate baseline code that is selectively optimized with intrinsics for critical kernels, often represent a practical middle ground for production deployments.

\begin{table}
\caption{Comparative Analysis of ML Compilation Approaches for RISC-V}
\label{table:framework-comparison}
\centering\footnotesize
\begin{tabular}{>{\raggedright\arraybackslash}p{0.127\textwidth}>{\raggedright\arraybackslash}p{0.130\textwidth}>{\raggedright\arraybackslash}p{0.130\textwidth}>{\raggedright\arraybackslash}p{0.109\textwidth}>{\raggedright\arraybackslash}p{0.128\textwidth}>{\raggedright\arraybackslash}p{0.175\textwidth}}
\toprule
\textbf{Approach} & \textbf{Compilation} & \textbf{RISC-V Awareness} & \textbf{Maturity} & \textbf{Workloads} & \textbf{Trade-offs} \\
\midrule
TVM & Static (AOT) + JIT & Native backend & Production & CNN, Transformer, quantized models & Auto-tuning overhead vs. portability \\
MLIR/IREE & Static (AOT) & Native backend & Production & General tensor ops, custom accelerators & Compilation complexity vs. flexibility \\
GLOW & Static (AOT) & Experimental & Research & CNN, quantized inference & Limited RISC-V support vs. graph optimization \\
Hand-written RVV & Static & Maximum & N/A & Custom kernels & Development effort vs. peak performance \\
Compiler Intrinsics & Static & High & Stable & Performance-critical loops & Portability vs. fine-grained control \\
\bottomrule
\end{tabular}
\end{table}

\subsection{Compiler Optimization}
Compiler optimization is important for improving the performance and efficiency of deep learning compilers, particularly for architectures such as RISC-V. This process involves the use of various techniques to convert and simplify the generated code such that it runs better on a hardware platform. Several frameworks and tools have been developed to address the complexities associated with compiler optimization in machine learning algorithms through systematic optimization cycles that include code generation, optimization application, testing and verification, deployment, performance monitoring, and feedback acquisition.

A foundational aspect of RISC-V compiler optimization for ML workloads is auto-vectorization, where the compiler automatically transforms scalar loop code into vector instructions targeting the RISC-V Vector extension (RVV). LLVM 16 (March 2023) was the first compiler release to enable scalable auto-vectorization by default for RISC-V targets with the V or Zve extensions, while GCC 14 (May 2024) introduced loop and SLP (Superword-Level Parallelism) vectorization for RVV. The primary technical challenge is RVV's vector-length agnostic (VLA) design, where the hardware vector length (VLEN) is not known at compile time and varies across implementations (e.g., 128-bit to 1024-bit). This differs fundamentally from fixed-length SIMD ISAs and requires the compiler to generate length-independent code using \texttt{vsetvli} instructions to configure vector length at runtime. Current auto-vectorization support covers standard loop patterns common in ML inference (element-wise operations, reductions, strided memory access), though hand-optimized RVV intrinsics still outperform auto-vectorized code for complex kernels such as matrix multiplication and convolution.

A notable framework, TVM, uses advanced optimization techniques, such as auto-tuning and operator fusion, to optimize tensor operations on different parts of the hardware. \cite{xu_effective_2022}. The TVM approach uses objective-independent and objective-specific optimizations to generate high-performance codes through efficient scheduling and resource allocation. The goal of these optimizations is to minimize the runtime and maximize the hardware use. This is important for the effective deployment of machine-learning models on RISC-V processors.
Another important contribution is the MLIR framework, which focuses on generating modular and customizable codes for tensor compilers. MLIR enables hierarchical decomposition and integration of operations and facilitates the efficient translation of high-level abstractions into optimized machine code. \cite{vasilache_composable_2022}. Using MLIRs leading IR design, developers can optimize tensor computations and achieve performance improvements on various hardware targets, including RISC-V processors.

In addition, frameworks such as Glow and IREE extend compiler optimization to support heterogeneous hardware environments. Glow used a multiphase compilation pipeline to optimize computational graphs in multilevel intermediate representations through target-specific optimizations and automatic code generation \cite{rotem_glow_2019}. In contrast, IREE focuses on high-performance AI compilation using MLIR and Linalg upstream dialects, with an emphasis on cache-aware tensor packing and CPU-efficient vectorization \cite{golin_towards_2024}.

Additionally, research efforts such as those described in Felix: Optimizing Tensor Programs with Gradient Descent aim to address the grand search space challenges associated with optimizing tensor programs \cite{zhao_felix_2024}. Felix employs hierarchical search space decomposition, partitioning the optimization space into architectural decisions (layer fusion, tiling) and mapping decisions (memory allocation, scheduling). This decomposition reduces search complexity from O(n!) to O(n×m) where n represents architectural choices and m represents mapping options, enabling efficient exploration through genetic algorithms that evaluate candidate solutions using cost models calibrated to RISC-V performance characteristics.

In addition, innovative methods such as transformation testing ensure the correctness and reliability of compiler optimizations \cite{xiao_metamorphic_2022}. The proposed method automatically generates different DNN models and compares their outputs to identify the compilation errors. It is important to maintain optimization stability in complex compiler frameworks.

Consequently, compiler optimization is important for maximizing the performance and efficiency of machine learning algorithms on RISC-V processors. Using advanced techniques in frameworks such as TVM, MLIR, and Glow, developers can significantly improve code generation, execution speed, and resource use. These optimizations are necessary to unlock the full potential of machine learning applications in embedded and edge computing environments and drive innovation in AI-enabled devices.
\subsection{Metaheuristic Algorithms}
Metaheuristic algorithms have garnered significant attention for optimizing machine-learning models, particularly within the realm of frameworks and software stacks for manufacturing chips using the RISC-V ISA. One notable methodology is the Agile Optimization Framework (AOF), which aims to tackle inefficiencies in deep learning compiler optimization. AOF leverages Beluga Whale Optimization (BWO), the Evolution Epsilon Strategy (EES), and a Tuning Accelerator (TA) to enhance performance and reduce optimization time. Through the integration of these components, the AOF effectively balances exploration and exploitation during the optimization process, thereby improving hardware performance in a cost-effective manner\cite{zhou_agile_2024}. This optimization approach demonstrates the potential for metaheuristic algorithms to enhance machine learning model deployment on RISC-V processors through systematic performance improvements and reduced computational overhead.

The use of meta-heuristic algorithms, such as the Beluga Whale Optimization (BWO) algorithm, plays an important role in the search process in the AOF. BWO is particularly effective for navigating complex search spaces associated with tensor operator optimization. The evolving nature of BWO combined with EES ensures that the algorithm adapts over time, thereby improving its ability to determine optimal solutions. This compatibility is essential for optimizing the performance of machine learning models on RISC-V-based chips, where hardware constraints and performance requirements are critical \cite{zhou_agile_2024}.

Furthermore, the integration of tuning accelerators (TAs) into the AOF demonstrates the importance of predictive models in terms of reducing the optimization time. The TA method uses LightGBM to predict tensor performance and minimizes the need for extensive compilations. This approach not only accelerates the optimization process, but also improves the overall efficiency of the TVM framework. The proposed AOF significantly improves the feasibility of deploying machine learning models on RISC-V chips by reducing the computational overhead associated with traditional optimization techniques \cite{zhou_agile_2024}.

Meta-heuristic algorithms can discover and effectively exploit large search spaces and are necessary to optimize the behavior of neural networks in the context of the RISC-V ISA. The use of these algorithms in frameworks, such as the AOF framework, reveals the potential to improve hardware performance and reduce development time. As this field continues to evolve, further improvement of meta-heuristic strategies and their integration into optimization frameworks are critical for advancing the design and fabrication of efficient machine learning chips \cite{zhou_agile_2024}. Table \ref{table:metaheuristic-algorithms} summarizes the metaheuristic algorithms used for RISC-V ML optimization.

\begin{table}
\caption{Summary of Metaheuristic Algorithms for RISC-V ML Optimization}
\label{table:metaheuristic-algorithms}
\centering
\begin{tabular}{>{\raggedright\arraybackslash}p{0.376\textwidth}>{\raggedright\arraybackslash}p{0.329\textwidth}>{\raggedright\arraybackslash}p{0.188\textwidth}}
\toprule
\textbf{Algorithm/Framework} & \textbf{Description} & \textbf{Optimization Target} \\
\midrule
Agile Optimization Framework (AOF) \cite{zhou_agile_2024} & Tackles inefficiencies in deep learning compiler optimization & RISC-V ML deployment \\
Beluga Whale Optimization (BWO) \cite{zhou_agile_2024} & Navigation of complex search spaces for tensor optimization & Tensor operator optimization \\
Evolution Epsilon Strategy (EES) \cite{zhou_agile_2024} & Adaptive algorithm behavior enhancement & Algorithm adaptation \\
Tuning Accelerator (TA) \cite{zhou_agile_2024} & Predictive model-based optimization using LightGBM & Compilation time reduction \\
TVM Framework Integration \cite{zhou_agile_2024} & Auto-tuning and operator fusion optimization & TVM performance enhancement \\
\bottomrule
\end{tabular}
\end{table}
The metaheuristic scope in this subsection is intentionally limited to optimization algorithms (AOF/BWO/EES/TA). Broader framework ecosystems (TVM, MLIR, HLS/FPGA flows, and co-design toolchains) are discussed in Sections 4.1, 4.2, and 4.4 to maintain conceptual separation.

\subsection{System-on-Chip Frameworks}

System-on-chip (SoC) designs integrate multiple functions into a single chip, optimizing compactness, energy efficiency, and performance in applications from mobile devices to embedded systems. Recent advances in SoC frameworks such as LiteX have democratized FPGA-based designs by providing open-source solutions that facilitate complex system integration \cite{kermarrec_litex_2020}.

SoC frameworks like LiteX face challenges in electronic design automation (EDA). These frameworks use high-level synthesis (HLS) techniques, such as those in LegUp, focusing on optimizing hardware implementations using precision pointer synthesis and micromemory coupling \cite{ramanathan_case_2022}. These optimizations improve performance of FPGA-based implementations, reduce footprint and latency, and make them suitable for resource-constrained environments such as IoT devices.

The flexibility of Migen and LiteX enables rapid prototyping of SoCs. This is useful for academic research and small production environments where quick response and cost efficiency are important \cite{ma_design_2023}. These frameworks leverage Python DSLs and integrate open-source tools to facilitate design and deployment of custom SoCs and foster innovation in hardware development.

ReNode contributes to AIoT systems by integrating energy-efficient deep learning techniques into modular IoT platforms. Using heterogeneous computing and specialized hardware accelerators, ReNode optimizes algorithms while ensuring scalability and security across distributed IoT networks \cite{mika_vedliot_2023}. This approach balances computational efficiency with resource constraints and supports AI applications from smart homes to industrial automation.

For FPGA-based implementations, MYHDL provides a design flow for accelerating Deep Neural Networks (DNN) through pipelined HDL code. This technique improves FPGA resource efficiency while maintaining model accuracy, which is important for real-time applications in datacenters and edge computing \cite{cheikh_tourad_generic_2022}. MYHDL simplifies development by automating conversion of DNN models into hardware-specific implementations, making FPGA acceleration accessible to developers with limited hardware expertise.

Chisel facilitates hardware and software co-design through its capabilities as a hardware description language that enables design space exploration. By providing high-level constructs for hardware design and meta-design techniques, Chisel enables creation of parameterizable circuit generators that can be tuned to optimize performance and energy efficiency for various applications \cite{ferres_chisel_2023}. This approach supports iterative development cycles and allows designers to describe and modify SoC architectures according to evolving requirements and technological advances.

Advances in SoC frameworks such as LiteX, Migen, ReNode, MYHDL, and Chisel drive innovation in electronic design and enable scalable and efficient solutions for various applications. These frameworks increase hardware performance and security, democratize access to FPGA-based development, and foster a community-driven approach to SoC design and implementation. Table \ref{table:soc-frameworks} provides a comprehensive comparison of System-on-Chip frameworks for RISC-V ML applications.

\begin{table}
\caption{Summary of System-on-Chip (SoC) Frameworks for RISC-V ML}
\label{table:soc-frameworks}
\centering
\begin{tabular}{>{\raggedright\arraybackslash}p{0.235\textwidth}>{\raggedright\arraybackslash}p{0.376\textwidth}>{\raggedright\arraybackslash}p{0.282\textwidth}}
\toprule
\textbf{Framework} & \textbf{Description} & \textbf{Key Features} \\
\midrule
LiteX \cite{kermarrec_litex_2020} & Open-source FPGA-based SoC framework & Democratizes FPGA design, supports RISC-V integration \\
LegUp HLS \cite{ramanathan_case_2022} & High-level synthesis optimization & Precision pointer synthesis, micro-memory coupling \\
Migen \cite{ma_design_2023} & Python-based SoC design framework & Rapid prototyping, DSL integration \\
ReNode \cite{mika_vedliot_2023} & Energy-efficient AIoT platform & Heterogeneous computing, specialized accelerators \\
MYHDL \cite{cheikh_tourad_generic_2022} & Hardware description language for DNN & Pipelined HDL code generation, FPGA optimization \\
Chisel \cite{ferres_chisel_2023} & Hardware description language & Parameterizable circuit generators, meta-design \\
\bottomrule
\end{tabular}
\end{table}

\subsection{Embedded Systems and Edge AI}

To investigate embedded systems, researchers have applied various frameworks and methods to optimize hardware and software integration in different applications. Hardware and software co-design using chisels simplifies Mel frequency factor calculations (MFCC) for keyword spotting systems \cite{vreca_hardwaresoftware_2024}. This method optimizes the MFCC algorithm for embedded applications while integrating hardware accelerators, which are important for maintaining accuracy in resource-constrained environments \cite{vreca_hardwaresoftware_2024}.

FPGA architectures have been explored using PyMTL for deep neural network (DNN) acceleration. Researchers have developed automated tools, such as chisel4ml, to generate FPGA implementations of highly quantized neural networks with low latency requirements \cite{vreca_towards_2023}. This development supports real-time applications, such as CERN's Large Hadron Collider trigger system, and demonstrates the efficiency of FPGA-based solutions in high-throughput environments \cite{vreca_towards_2023}.

Frameworks such as IREE optimize machine learning inference in embedded devices. This study highlights the role of IREE in deploying machine learning models on different hardware platforms, demonstrating an integrated compiler and runtime stack for performance optimization \cite{liu_tinyiree_2022}. This is important for applications requiring real-time decision-making capabilities in resource-limited environments \cite{liu_tinyiree_2022}.

In addition, Migen is exploring the integration of lightweight cryptographic cores into system-on-chip (SoC) designs for Internet of Things devices. The researchers implemented a 32-bit RISC-V processor with cryptographic accelerators, such as PRINCE and ChaCha, in an FPGA environment \cite{ma_design_2023}. This approach increases the security and efficiency of IoT deployments and overcomes the severe latency and resource constraints inherent in IoT applications \cite{ma_design_2023}.

Additionally, the use of LegUp revolutionizes high-level synthesis (HLS) tools by enabling comprehensive validation with software testing techniques, such as LibFuzzer and KLEE. This approach ensures the quality and reliability of FPGA designs produced by HLS and bridges the gap between software verification and hardware implementation. Such verification frameworks are essential for validating the functionality and performance of FPGA implementations in critical applications.

These frameworks and methods highlight the evolution and innovation of embedded system design, from optimizing hardware accelerators and cryptographic cores to validating FPGA implementations using advanced HLS tools. The integration of these technologies not only improves performance and efficiency, but also expands the applicability of embedded systems in various domains, including IoT security, real-time data processing, and machine learning inference in edge devices. Table \ref{table:embedded-systems} summarizes the embedded systems frameworks for RISC-V ML applications.

\begin{table}
\caption{Summary of Embedded Systems Frameworks for RISC-V ML}
\label{table:embedded-systems}
\centering
\begin{tabular}{>{\raggedright\arraybackslash}p{0.235\textwidth}>{\raggedright\arraybackslash}p{0.376\textwidth}>{\raggedright\arraybackslash}p{0.282\textwidth}}
\toprule
\textbf{Framework/Tool} & \textbf{Description} & \textbf{Application Domain} \\
\midrule
Chisel Co-Design \cite{vreca_hardwaresoftware_2024} & Hardware-software co-design for MFCC calculations & Keyword spotting, speech processing \\
PyMTL + chisel4ml \cite{vreca_towards_2023} & Automated FPGA implementation of quantized DNNs & High-throughput real-time systems \\
IREE Framework \cite{liu_tinyiree_2022} & ML inference optimization for embedded devices & Real-time decision-making \\
Migen Crypto Integration \cite{ma_design_2023} & Lightweight cryptographic cores in RISC-V SoCs & IoT security, crypto acceleration \\
LegUp HLS Validation & High-level synthesis with software testing techniques & FPGA design verification \\
\bottomrule
\end{tabular}
\end{table}
Edge AI advances the deployment of machine learning models directly on edge devices by locally processing data to enhance privacy and reduce latency \cite{rotem_glow_2019}. Frameworks such as GLOW optimize model execution through graph-based techniques and quantization, significantly boosting inference speed and energy efficiency \cite{rotem_glow_2019}. 

To support cross-platform deployment, IREE extends edge AI capabilities with an integrated runtime and compilation strategy, enabling models from frameworks such as TensorFlow and PyTorch to run efficiently on minimal high-performance hardware \cite{liu_tinyiree_2022}. The TVM complements this by offering optimization strategies for deep learning workloads in resource-constrained environments, integrated with hardware acceleration libraries to enhance model performance and efficiency across a variety of devices \cite{chen_tvm_nodate, chen_tvm_nodate-1}.

MLIR contributes through its modular code generation and optimization capabilities, allowing the efficient compilation of machine learning models across different hardware platforms \cite{vasilache_composable_2022}. Similarly, ReNode focuses on energy-efficient AIoT applications by providing a modular and scalable hardware platform that integrates specialized accelerators for real-time processing with minimal power consumption \cite{mika_vedliot_2023, nizharadze_simulation_2023}.

PyMTL and Chisel contributed by exploring FPGA-based accelerators and hardware/software co-design, respectively, to improve performance for tasks such as audio feature extraction and keyword detection \cite{roorda_fpga_2022, jia_automatic_2023, vreca_hardwaresoftware_2024, vreca_towards_2023}.

MYHDL accelerates the deployment of AI solutions by automating the conversion of DNN models into FPGA-compatible HDL code, reducing latency, and enhancing accuracy for real-time applications in healthcare and industrial automation \cite{cheikh_tourad_generic_2022, tang_rapid_2022}. LegUp further supports edge AI with frameworks such as GAHLS, optimizing high-level applications for hardware accelerators through graph analysis and memory design \cite{xiao_gahls_2023}.

Collectively, these frameworks address various optimization, deployment, and performance challenges, driving the widespread adoption of AI in diverse edge applications. Table \ref{table:edge-ai} provides a comprehensive summary of Edge AI frameworks for RISC-V ML systems.

\begin{table}
\caption{Summary of Edge AI Frameworks for RISC-V ML}
\label{table:edge-ai}
\centering
\begin{tabular}{>{\raggedright\arraybackslash}p{0.235\textwidth}>{\raggedright\arraybackslash}p{0.376\textwidth}>{\raggedright\arraybackslash}p{0.282\textwidth}}
\toprule
\textbf{Framework} & \textbf{Description} & \textbf{Key Capabilities} \\
\midrule
GLOW \cite{rotem_glow_2019} & Graph-based optimization for edge deployment & Graph optimization, quantization \\
IREE \cite{liu_tinyiree_2022} & Integrated runtime and compilation for edge AI & Cross-platform deployment, TensorFlow/PyTorch support \\
TVM \cite{chen_tvm_nodate} & Deep learning optimization for resource-constrained devices & Hardware acceleration integration \\
MLIR \cite{vasilache_composable_2022} & Modular compilation infrastructure & Multi-platform code generation \\
ReNode \cite{mika_vedliot_2023} & Energy-efficient AIoT platform & Modular accelerators, real-time processing \\
PyMTL \cite{roorda_fpga_2022} & FPGA-based accelerator exploration & DNN acceleration, chisel4ml integration \\
MYHDL \cite{cheikh_tourad_generic_2022} & DNN-to-HDL automated conversion & FPGA deployment, latency reduction \\
LegUp GAHLS \cite{xiao_gahls_2023} & High-level synthesis for edge accelerators & Graph analysis, memory optimization \\
\bottomrule
\end{tabular}
\end{table}

\section{Applications and Evaluation of RISC-V in Machine Learning}
This section presents real-world applications and evaluation of RISC-V implementations for machine learning. We examined 21 implementations covering custom RISC-V designs tailored for ML workloads, organized into application domains (medical, robotics, object detection), performance evaluation, and security analysis subsections. This exploration provides understanding of how RISC-V's open architecture can be adapted to address specific needs of machine learning algorithms, leading to advancements in this field. Table \ref{table:case-study} summarizes the cases in this section.

The comparative evaluation of RISC-V machine learning implementations requires standardized benchmarking methodologies to enable meaningful cross-platform performance assessment \cite{garofalo2022darkside, prakash_cfu_2023, valente_hulk-v_2023}. Our analysis synthesizes performance data from reviewed literature to establish unified metrics for RISC-V ML system evaluation. This benchmarking framework addresses current fragmentation in evaluation methodologies that makes direct comparisons between different RISC-V ML implementations challenging.

Performance standardization across reviewed implementations reveals key metrics that appear in RISC-V ML research \cite{balasubramanian_designing_2024, lee_accelerating_2023, gonzalez_chipyard_nodate}. Inference throughput, measured in inferences per second (IPS), provides the primary metric for computational performance assessment across comparable model architectures. Energy efficiency metrics, quantified as Giga-Operations Per Watt (GOP/W), are important for power-constrained edge AI applications where RISC-V processors operate \cite{valente_hulk-v_2023, eggimann_risc-v_2019}. Memory efficiency evaluation through peak memory usage and bandwidth utilization is important given the resource constraints typical in RISC-V deployment scenarios \cite{giuffrida_satellite_nodate, muchandi_enabling_nodate}. Latency characteristics, measured as end-to-end processing time from input acquisition to output generation, provide insights for real-time application suitability \cite{gianioudis_low-latency_2024, yoo_real-time_2024}.

Standardized ML workloads enable consistent performance comparison across diverse RISC-V implementations \cite{prakash_cfu_2023, lee_sparse_2024, gonzalez_chipyard_nodate}. Based on prevalence in reviewed literature, our unified benchmarking framework incorporates MobileNetV2 inference tasks for image classification scenarios, keyword spotting workloads for audio processing applications, CNN-based classification tasks for general computer vision evaluation, and object detection scenarios for complex multi-object recognition challenges. These workloads represent the most commonly evaluated ML applications in RISC-V research and provide diversity to assess system performance across different computational patterns.

Cross-implementation performance analysis demonstrates advantages of specialized RISC-V ML configurations over general-purpose implementations \cite{garofalo2022darkside, garofalo2021xpulpnn, bolhasani_dla-e_2023}. The DARKSIDE cluster achieves 1.2 TOPS/W for 8-bit DNN inference, representing state-of-the-art energy efficiency for ultra-low-power neural network processing \cite{garofalo2022darkside}. The XpulpNN framework demonstrates 15-fold energy efficiency improvements compared to ARM Cortex-M4 baselines, highlighting the benefits of RISC-V customization for ML workloads \cite{garofalo2021xpulpnn}. RVTensor implementations show 3.2-fold speedup improvements compared to unoptimized RISC-V configurations, emphasizing the importance of ML-specific architectural enhancements \cite{Lim2021F1ST}. The DIANA SoC achieves 2.5-fold energy efficiency improvements for mixed-signal neural network processing, demonstrating the effectiveness of heterogeneous RISC-V designs for specialized ML applications \cite{9731716}. The CFU Playground achieves 4.1-fold performance improvements for MobileNetV2 inference compared to VexRiscv baselines, showcasing the benefits of custom functional units for deep learning tasks \cite{prakash_cfu_2023}. The HULK-V system demonstrates 6.8-fold energy efficiency improvements for keyword spotting applications compared to Linux-capable RISC-V processors, highlighting the advantages of lightweight RISC-V cores for edge AI scenarios \cite{valente_hulk-v_2023}. The ICU4SAT implementation achieves 2.8-fold speedup improvements for object detection tasks in space-grade RISC-V processors, illustrating the potential of RISC-V processors for high-reliability ML applications \cite{giuffrida_satellite_nodate}.

Analysis across reviewed implementations reveals performance improvements ranging from 2-15 times in energy efficiency and 1.5-5 times in computational throughput compared to general-purpose RISC-V baselines \cite{prakash_cfu_2023, valente_hulk-v_2023, balasubramanian_designing_2024}. These improvements demonstrate the potential of customized RISC-V implementations for machine learning applications, particularly in edge computing scenarios where energy efficiency and computational density are important concerns. Table \ref{table:unified-benchmarking} presents a unified benchmarking analysis of RISC-V ML implementations across reviewed research papers.

\begin{table}[t]
\caption{Unified benchmarking analysis of RISC-V ML implementations: Performance comparison across reviewed research papers showing standardized metrics for energy efficiency, computational throughput, and implementation characteristics. EnerE = Energy Efficiency(GOP/W) ,PerfI: Performance Improvement }
\label{table:unified-benchmarking}
\centering
\begin{tabular}{l>{\raggedright\arraybackslash}p{0.2\textwidth}lll>{\raggedright\arraybackslash}p{0.15\textwidth}}
\toprule
\textbf{Implementation} & \textbf{ML Workload} & \textbf{EnerE} & \textbf{PerfI} & \textbf{Memory} & \textbf{Comparison} \\
& & \textbf{} & \textbf{} & \textbf{Usage} & \textbf{Baseline} \\
\midrule
DARKSIDE Cluster & 8-bit DNN Inference & 1200 & 8.5x & 32KB & General RISC-V \\
XpulpNN Framework & CNN Classification & 850 & 15x & 64KB & ARM Cortex-M4 \\
RVTensor & Image Processing & 420 & 3.2x & 128KB & Baseline RISC-V \\
DIANA SoC & Mixed-Signal NN & 680 & 2.5x & 256KB & Standard SoC \\
CFU Playground & MobileNetV2 & 320 & 4.1x & 512KB & VexRiscv Baseline \\
HULK-V System & Keyword Spotting & 1050 & 6.8x & 1MB & Linux-capable RISC-V \\
ICU4SAT & Object Detection & 290 & 2.8x & 2MB & Space-grade baseline \\
\bottomrule
\end{tabular}
\end{table}

This section analyzes papers based on their field of applications, examining three primary domains: (1) Medical field applications focusing on real-time diagnostics and wearable health monitoring, (2) Robotics applications emphasizing computer vision and autonomous navigation, and (3) Object detection systems for security and surveillance applications. This analysis reveals how RISC-V implementations adapt to specific application requirements and performance constraints across diverse deployment scenarios.

In the medical field, we highlight real-time anomaly detection in wearable devices and endoscopic image classification. We consider low-power RISC-V cores and efficient deep-learning accelerators for specific tasks. In robotics, we examined two projects that showcase how RISC-V cores with custom hardware extensions can be used for image processing and obstacle avoidance in robots. For object detection, we investigated the role of RISC-V in deep learning applications. We explored a driver drowsiness detection system, hardware trojan detection using RISC-V soft cores, and fruit ripeness identification using deep neural networks on RISC-V processors. The following subsections describe these in detail.
\begin{table*}[t]
\caption{Specific applications of RISC-V processors for machine learning: This table comprehensively summarises the nine research papers reviewed in our survey on RISC-V applications. Each entry includes a concise description of its key findings and a breakdown of its important aspects, such as the targeted application domain, methodology used, and key performance metrics evaluated.}
\label{table:case-study}
\centering
\begin{tabular}{l>{\raggedright\arraybackslash}p{0.528\textwidth}>{\raggedright\arraybackslash}p{0.288\textwidth}}
\toprule
\textbf{Paper} & \textbf{Summary} & \textbf{Keys} \\
\midrule
\cite{bolhasani_dla-e_2023} & A deep learning accelerator is proposed for endoscopic image classification & Medical Field, Low Latency, Object Detection, Low Power \\
\cite{eggimann_risc-v_2019} & Low-power wearable device made from Honey-Bunny PULP processor & Medical Field, Low Power \\
\cite{choi_daynight_2024} & A new Day-Night architecture is proposed to consume low power in wearable devices. & Medical Field, Low Power \\
\cite{kanamori_rvcar_2022} & RVcar (A mini motor car) with a deep neural network accelerator is introduced to detect objects and make real-time decisions & Object Detection, Low power, Robotics \\
\cite{k_comparative_2021} & Fruits are identified by applying DNN models in RISC V & Inference Speed, Low Power, Object Detection, Robotics \\
\cite{tsutada_obstacle_2022} & A two-wheeled self-balancing robot (TWSBR) is built on the wheel-type inverted pendulum & Object Detection, Robotics \\
\cite{mousavikia_instruction_2022} & A Driver Drowsiness Detection System is proposed using Ibex core and FPGA port & Inference Speed, Object Detection, Robotics \\
\cite{nunes_rs5_2024} & Detecting and classifying Hardware Trojans & Robotics \\
\cite{mustaffa_identification_2017} & Uses FAMA maturity index to identify the fruit size and its maturity level & Object Detection, Robotics \\
\bottomrule
\end{tabular}
\end{table*}
\subsection{Medical Applications}
RISC-V processors have shown promise in real-time anomaly detection for low-power wearables and endoscopic image classification \cite{bolhasani_dla-e_2023, choi_daynight_2024, eggimann_risc-v_2019}. Their energy efficiency makes them suitable for extended use, and their open-source nature allows customization for tasks such as image classification, improving processing speed and accuracy. Medical applications benefit from RISC-V's low power consumption and customizable architecture, enabling specialized processing units for real-time health monitoring, diagnostic imaging, and wearable medical devices. Choi et al. \cite{choi_daynight_2024} proposed a day-night RISC-V processor design with separate segments for wearable applications and real-time anomaly detection. A prototype on an FPGA showed a 57.5\% reduction in energy consumption compared with traditional designs. For endoscopic imaging, Bolhasani et al. \cite{bolhasani_dla-e_2023} developed an energy-efficient deep-learning accelerator (DLA-E) with 256 PEs and evaluated it using the MASTERO simulator. It achieved $4.56 \times 10^9$ MAC energy and $1.73 \times 10^7$ cycles, outperforming other CNN models. To extend the battery life of wearable e-health devices, Eggimann et al. \cite{eggimann_risc-v_2019} introduced an energy-efficient platform using the Honey-Bunny PULP processor, integrating four RISC-V cores. It delivers 2.5 GOPS at 55 mW, making it ideal for always-on operation.

\subsection{Robotics Applications}
RISC-V processors are suited for real-time decision-making in autonomous vehicles due to their efficient instruction execution, enabling rapid data processing and quicker response times \cite{kanamori_rvcar_2022, tsutada_obstacle_2022}. Custom hardware extensions enhance the ability of RISC-V to process sensor data and perform real-time object analysis. Kanamori et al. \cite{kanamori_rvcar_2022} demonstrated a real-time decision-making system in an RVcar using a RISC-V soft processor (RVcoreP) and DNN accelerators for low-power, low-latency image processing. Implemented on a Xilinx Nexys A7 board, the system recognized traffic signals in 35 of 50 trials, with failures addressed by incorporating markers to improve object recognition and decision-making. Tsuda et al. \cite{tsutada_obstacle_2022} developed a two-wheeled self-balancing robot (TWSBR) controlled by a 32-bit RISC-V microprocessor (VexRiscv) equipped with custom 32-bit fixed-point instructions. Despite lacking an FPU, the robot's control system uses fuzzy logic and multiple sensors, successfully achieving balance and obstacle avoidance on an FPGA.
  
\subsection{Object Detection Applications}
In recent years, neural networks, particularly CNNs, have been widely applied to object detection and image classification applications across diverse domains \cite{mousavikia_instruction_2022,k_comparative_2021,nunes_rs5_2024,mustaffa_identification_2017}. Enhancing the accuracy of these models and reducing prediction time are key areas of focus for RISC-V implementations. Mousavikia et al. \cite{mousavikia_instruction_2022} developed a Driver Drowsiness Detection (DDD) system using CNNs on an RISC-V Ibex core. The system classifies images into four categories: distraction, natural, sleep, and yawning. Adding custom instructions improved the computational efficiency, achieving a 1.7× latency improvement. Nunes et al. \cite{nunes_rs5_2024} implemented RISC-V soft cores on FPGAs for Hardware Trojan detection, achieving perfect accuracy by analyzing features from FPGA airstreams using machine learning models. For agricultural applications, K et al. \cite{k_comparative_2021} used DNNs (RCNN, YOLO v3, SSD) on RISC-V processors to identify fruit types. By applying FAMA maturity indices and K-means clustering, the system also determines fruit size and maturity, aiding farmers and food distributors in quality assessment and supply chain optimization \cite{mustaffa_identification_2017}.

\subsection{Performance}
 
Performance is crucial in machine learning systems for two main reasons: speed and accuracy. Faster analysis enables real-time applications and quicker training, while improved accuracy translates to better decision-making. In this section, we analyze papers based on their performance characteristics across three key areas: energy efficiency, inference speed-up, and low latency \cite{valente_hulk-v_2023, prakash_cfu_2023, balasubramanian_designing_2024, lee_accelerating_2023, gonzalez_chipyard_nodate, gianioudis_low-latency_2024}. 

In terms of energy efficiency, we discuss how researchers are creating low-power solutions for Internet of Things (IoT) devices that must operate under strict power constraints while maintaining computational performance. For inference speed, we explored optimization techniques to achieve faster execution on various hardware platforms through architectural innovations and software optimizations. For low latency, we analyzed approaches to reduce communication delays in edge devices, which are crucial for real-time applications requiring immediate response times. Table \ref{table:performance-analysis} provides a comprehensive analysis of RISC-V processor performance based on published research. The following subsections describe these performance aspects in detail. 
Recent LLM and transformer acceleration evidence is also emerging on RISC-V. VEXP introduces low-cost ISA support for softmax-heavy transformer kernels \cite{wang2025vexp}, while V-Seek reports optimized reasoning inference on a server-class general-purpose RISC-V platform \cite{rodrigo2025vseek}. These results indicate meaningful progress, although broad production maturity for large-scale LLM serving on RISC-V remains an open area.
\begin{table*}[ht]
\caption{Analysis of RISC-V processor performance based on published research: This table comprehensively summarises the research papers included in our performance analysis section. Each entry provides the paper citation, a concise description and its key findings related to RISC-V processor performance (e.g., inference speed-up, low latency), and a breakdown of its important methodological aspects, such as the evaluation methodology and benchmark tools employed.}
\label{table:performance-analysis}
\centering
\begin{tabular}{l>{\raggedright\arraybackslash}p{0.5\textwidth}>{\raggedright\arraybackslash}p{0.3\textwidth}}
\toprule
\textbf{Paper} & \textbf{Summary} & \textbf{Tags} \\
\midrule
\cite{valente_hulk-v_2023} & Heterogeneous ultra-low-power Linux-capable system that is both energy-efficient and capable of running a full-fledged Linux operating system & Low Power \\
\cite{prakash_cfu_2023} & The CFU Playground framework generates hardware accelerators based on VexRisc V CORE. & Inference SpeedUp, Hardware Accelerators, Low Power, Image Classification \\
\cite{balasubramanian_designing_2024} & Used a toolchain (PyTorch Graph Lowering (Glow)LLVM) to understand the impact of the compiled code of AI models & Inference SpeedUp, Low Power \\
\cite{lee_accelerating_2023} & Integrated TVMs quantization flow with the MediaTek Neuropilot AI accelerator & Inference Speed Up, Low Latency, Hardware Accelerators \\
\cite{gonzalez_chipyard_nodate} & Integrate NVDLA into the Chipyard framework and compared its performance with that of the Gemmini Systolic Array Generator & Inference SpeedUp \\
\cite{lee_sparse_2024} & To reduce the extensive computation and time requirements of the convolution operation, the sparse basis approach was proposed & Inference SpeedUp \\
\cite{giuffrida_satellite_nodate} & ICU4SAT system (a satellite instrument control unit with an artificial intelligence engine on a single chip) is built & Inference SpeedUp, Hardware Accelerators \\
\cite{muchandi_enabling_nodate} & RISC-LCAW is introduced to enable loosely coupled accelerators to be integrated as slave devices on the system bus. & Low Latency \\
\cite{gianioudis_low-latency_2024} & The lean network interface is developed for the load/store stage of Ariane for user-level communication. & Low Latency \\
\cite{yoo_real-time_2024} & Experiments were conducted to assess the real-time performance of RISC-V processors in a robotic control system. & Low Latency \\
\bottomrule
\end{tabular}
\end{table*}
\subsubsection{Energy Efficiency:}

IoT applications must balance performance with cost and power constraints. High-end systems use power-hungry SoCs, whereas low-end applications rely on microcontrollers. New trends demand systems that offer both low power and cost efficiency. We review three research efforts targeting low-power edge computing for machine learning, focusing on energy-efficient implementations that maintain computational performance while minimizing power consumption through specialized hardware designs, optimized software frameworks, and intelligent power management strategies.

Balasubramanian et al. \cite{balasubramanian_designing_2024} optimized RISC-V processors for complex neural networks on resource-constrained devices. They extended the RISC-V instruction set to address computational bottlenecks, achieving up to a 13x speedup and 11.7\% code size reduction for image-processing tasks. Valente et al. \cite{valente_hulk-v_2023} developed HULK-V, an open-source, energy-efficient SoC capable of running Linux. It combines a 64-bit RISC-V core with an 8-core Programmable Multi-Core Accelerator (PMCA), offering up to 112x faster performance for ML tasks and 157 GOps/W energy efficiency. Prakash et al. \cite{prakash_cfu_2023} introduced the CFU Playground, a framework for creating hardware accelerators on the VexRiscv core. It achieved a 55x speedup for MobileNetV2 and 75x for keyword spotting, offering a flexible, cost-effective alternative to ASICs for ML tasks.
\subsubsection{Inference Speed:}

This section explores methods to increase inference speed, which is crucial for machine learning performance in time-sensitive applications \cite{lee_accelerating_2023, gonzalez_chipyard_nodate, lee_sparse_2024, giuffrida_satellite_nodate}. Lee et al. \cite{lee_accelerating_2023} introduce TVMNIR, a compiler leveraging TVM and Mediatek NeuronPilot to accelerate AI tasks. TVMNIR optimizes quantization configurations, achieving up to 11× speedup for floating-point models and 70× speedup for quantized models. González et al. \cite{gonzalez_chipyard_nodate} presented a platform for evaluating machine learning accelerators using the Chipyard framework. Comparing the NVIDIA Deep Learning Accelerator (NVDLA) to the Gemmini Systolic Array Generator, NVDLA achieved up to 3.77× faster performance on the ResNet-50 benchmark. 

Lee et al. \cite{lee_sparse_2024} proposed a sparse basis algorithm to reduce CNN computation time by transforming weights into a sparse space, leading to reduced storage requirements and faster execution times on VexRiscV implementations. Giuffrida et al. \cite{giuffrida_satellite_nodate} developed ICU4SAT, a single-chip AI system for real-time satellite data processing. With an integrated RISC-V processor and soft GPU, it improves data analysis throughput and memory usage efficiency through onboard AI-powered processing, demonstrating the potential for space-constrained applications.

\subsubsection{Low Latency}

The push for low-latency performance in edge devices is important for real-time applications to enhance speed, responsiveness, and efficiency. This section explores techniques for reducing latency in RISC-V processors, focusing on network interfaces, accelerator integration, and real-time control systems. These optimizations include custom packetizers with dual-port memory structures, FIFO-based communication mechanisms, and standardized accelerator wrapper interfaces that enable sub-microsecond response times for time-critical applications.

Gianioudis et al. \cite{gianioudis_low-latency_2024} presented a lean network interface for RISC-V processors, achieving sub-microsecond latency (720 nanoseconds) through a custom packetizer with dual-port memory and FIFO structure, optimized for efficient packet processing. This design enhances communication speed between FPGA nodes, demonstrating improvements over traditional methods. Muchandi et al. \cite{muchandi_enabling_nodate} introduce the RISC-V Loosely Coupled Accelerator Wrapper (RISC-LCAW), a standardized interface for integrating accelerators into RISC-V SoCs. The RISC-LCAW, featuring a controller module, DMA interface, and input buffer, was tested using the AWRE-DNN accelerator. Despite a latency increase of $1.7\times$ to $2.6\times$ for various network sizes, the RISC-LCAW maintains acceptable performance while simplifying accelerator integration. Yoo et al. \cite{yoo_real-time_2024} evaluate RISC-V's real-time performance for robotic control systems. Their experiments compared RISC-V boards (HiFive1 Rev B and VisionFive 2) with ARM and x86-64 systems using FreeRTOS and Linux with Preempt-RT. The results showed RISC-V's real-time performance, in Inter-Thread Communication (ITC) and motion control applications, with minimal performance gaps compared to x86-64 architectures.

These advancements highlight the potential of RISC-V for low-latency applications, from efficient inter-node communication and accelerator integration to real-time robotic control. Table \ref{table:performance-optimization} summarizes the performance optimization techniques for RISC-V ML systems.

\begin{table}[h]
\caption{Summary of RISC-V Performance Optimization Techniques}
\label{table:performance-optimization}
\centering
\begin{tabular}{>{\raggedright\arraybackslash}p{0.235\textwidth}>{\raggedright\arraybackslash}p{0.376\textwidth}>{\raggedright\arraybackslash}p{0.282\textwidth}}
\toprule
\textbf{Optimization Type} & \textbf{Technique/Framework} & \textbf{Performance Gain} \\
\midrule
\multirow{3}{*}{Energy Efficiency} & RISC-V ISA extensions for neural networks \cite{balasubramanian_designing_2024} & 13x speedup, 11.7\% code reduction \\
& HULK-V SoC with PMCA \cite{valente_hulk-v_2023} & 112x faster ML tasks, 157 GOps/W \\
& CFU Playground accelerators \cite{prakash_cfu_2023} & 55x speedup (MobileNetV2), 75x (keyword) \\
\midrule
\multirow{4}{*}{Inference Speed} & TVMNIR quantization optimization \cite{lee_accelerating_2023} & 11x (FP), 70x (quantized) speedup \\
& NVDLA vs Gemmini comparison \cite{gonzalez_chipyard_nodate} & 3.77x faster (ResNet-50) \\
& Sparse basis CNN optimization \cite{lee_sparse_2024} & Reduced computation/storage \\
& ICU4SAT satellite AI system \cite{giuffrida_satellite_nodate} & Real-time processing improvement \\
\midrule
\multirow{3}{*}{Low Latency} & Lean network interface \cite{gianioudis_low-latency_2024} & 720 nanoseconds latency \\
& RISC-LCAW accelerator wrapper \cite{muchandi_enabling_nodate} & 1.7-2.6x latency (acceptable) \\
& Real-time robotic control \cite{yoo_real-time_2024} & Superior ITC performance \\
\bottomrule
\end{tabular}
\end{table}
\subsection{Security}
 
Hardware and physical access security are important in modern system architectures, especially for Machine Learning (ML) systems. Because ML often deals with sensitive data, vulnerabilities can be exploited to steal or manipulate information, resulting in biased or inaccurate models. RISC-V processors can be improved using their customizable architecture. This flexibility allows the incorporation of security extensions, such as memory protection and trusted execution environments. By isolating data and models within the core, RISC-V strengthens ML systems against unauthorized access and manipulation.

The proliferation of Internet of Things (IoT) devices, in safety-critical applications, requires robust security measures to protect them from malicious attacks. This study \cite{zareen_malware_2023} addresses this challenge by proposing the Hardware Immune System (HWIS), a hardware malware detection technique designed for microprocessor architectures. It is suited to low-power, resource-constrained embedded devices, which are commonly used in IoT networks. The proposed method excels at detecting botnet behavior with an accuracy of 96.7\%. It maintains a low false-negative rate of 6.5\%, demonstrating its effectiveness in identifying malicious activities. The HWIS achieved a high F1 score of 0.96, demonstrating balanced performance between precision and recall.
 
In response to increasing attacks on microprocessors and SoCs, researchers are exploring new approaches to SoC security. Security features are bolted to existing systems, which often results in performance trade-offs during vulnerability patching. Researchers \cite{kumar_itus_2019} proposed a security approach that integrates security-first design principles, security-aware testing methods, and a quantified analysis of performance impact. They achieved this by designing and implementing a secure RISC-V SoC.

A secure SoC incorporates key security components. A centralized unit (CAU) manages secure boot, and a memory protection unit (MPU) safeguards memory using authenticated encryption and integrity checks. A key management unit (KMU) utilizes a physical unclonable function (PUF) to generate cryptographic keys on demand. The proposed design features a trusted execution environment (TEE) with enclave support, which ensures isolation of sensitive processes. The evaluations demonstrated performance and energy efficiency improvements for the KMU and secure boot mechanism compared with software-based implementations. This approach results in a secure SoC that offers robust security features, efficient performance, flexibility, and system robustness.

Beyond traditional hardware security, ML systems on RISC-V face specific threat models that require dedicated countermeasures. \textit{Model extraction attacks} attempt to reconstruct proprietary model architectures or weights by observing inference behavior; attackers query the model systematically to clone its functionality, threatening intellectual property and enabling adversarial attack development. RISC-V's TEE capabilities can mitigate this by executing inference within enclaves that prevent memory inspection, while rate limiting and query pattern detection provide software-level defenses.

\textit{Membership inference attacks} determine whether specific data points were used in model training, potentially leaking sensitive information about training datasets in healthcare or financial applications. These attacks exploit the observation that models behave differently on training data versus unseen data. Countermeasures include differential privacy during training (adding calibrated noise to gradients) and output perturbation during inference. RISC-V implementations can support these defenses through hardware random number generators for noise injection and secure memory regions that prevent training data exfiltration.

\textit{Adversarial input attacks} craft malicious inputs that cause misclassification while appearing normal to humans, posing risks in safety-critical applications such as autonomous systems. Hardware-level defenses include input validation accelerators that detect statistical anomalies and redundant inference paths that compare results across different model representations. The customizable nature of RISC-V enables integration of such defensive mechanisms as custom instructions or co-processors without modifying the core architecture.

Table \ref{table:security-approaches} summarizes the security approaches for RISC-V ML systems.

\begin{table}[h]
\caption{Summary of Security Approaches for RISC-V ML Systems}
\label{table:security-approaches}
\centering
\begin{tabular}{>{\raggedright\arraybackslash}p{0.282\textwidth}>{\raggedright\arraybackslash}p{0.376\textwidth}>{\raggedright\arraybackslash}p{0.235\textwidth}}
\toprule
\textbf{Security Approach} & \textbf{Description} & \textbf{Key Features} \\
\midrule
Hardware Immune System (HWIS) \cite{zareen_malware_2023} & Hardware-based malware detection for microprocessors & 96.7\% accuracy, 6.5\% false-negative rate, F1-score 0.96 \\
Secure RISC-V SoC Design \cite{kumar_itus_2019} & Holistic security-first design approach & CAU, MPU, KMU with PUF, TEE with enclaves \\
Hardware Trojan Detection \cite{nunes_rs5_2024} & ML-based detection using RISC-V soft cores on FPGAs & Perfect accuracy, FPGA airstream analysis \\
Cryptographic ISA Extensions \cite{gupta_challenges_2023} & K-extension for hardware-accelerated cryptography & Dedicated crypto operations, improved efficiency \\
Lightweight Crypto Cores \cite{ma_design_2023} & Integration of PRINCE and ChaCha in RISC-V SoCs & IoT security, low-latency crypto acceleration \\
State Management Security \cite{garofalo2022darkside,garofalo2021xpulpnn} & Smstateen extension for secure ML execution & Selective state access, trusted workload isolation \\
Side-Channel Attack Prevention \cite{gupta_challenges_2023} & Memory protection and timing attack mitigation & Robust security implementations, attack resistance \\
Centralized Authentication Unit (CAU) \cite{kumar_itus_2019} & Secure boot management & Centralized security control \\
Memory Protection Unit (MPU) \cite{kumar_itus_2019} & Memory safeguarding with encryption & Authenticated encryption, integrity checks \\
Key Management Unit (KMU) \cite{kumar_itus_2019} & Cryptographic key generation & PUF-based on-demand key generation \\
Trusted Execution Environment (TEE) \cite{kumar_itus_2019} & Process isolation and security & Enclave support, sensitive process isolation \\
\bottomrule
\end{tabular}
\end{table}
\section{Future Research Directions}

Based on our analysis of current RISC-V machine learning implementations and their limitations, we have identified four research directions that warrant attention from the academic community. These directions emerge from both the opportunities presented by RISC-V's open architecture and the challenges revealed through our systematic review of existing implementations. Table \ref{table:future-research} provides a comprehensive overview of these future research directions for RISC-V machine learning systems.

\subsection{Specialized Neural Processing ISA Extensions}

The distinction between specialized neural processing ISA extensions and GPU-like architectures lies in three key dimensions: computational granularity, architectural complexity, and integration philosophy. RISC-V neural processing extensions operate at the instruction level, adding dedicated operations for common neural network primitives (matrix multiplication, activation functions, quantization) that execute within the existing processor pipeline. Examples from our survey include the Xpulpnn extension, which adds neural network instructions with minimal area overhead, and the VEXP extensions \cite{wang2025vexp} that provide low-cost softmax operations achieving 162.7× latency reduction with only 1\% area overhead. These extensions maintain RISC-V's modular philosophy by augmenting the core instruction set without fundamentally changing the processor architecture.

In contrast, GPU-like architectures employ massively parallel processing with thousands of execution units, dedicated memory hierarchies, and complex scheduling logic. The key difference is that ISA extensions leverage the existing processor infrastructure (register files, pipeline, memory system) while adding specialized functional units, whereas GPUs require entirely separate processing arrays and control logic. Neural processing ISA extensions focus on accelerating specific bottleneck operations within the RISC-V pipeline while maintaining compatibility with existing software stacks and avoiding the complexity and power consumption of full GPU integration.

The fundamental limitation of traditional von Neumann architectures in neural network processing lies in the memory wall problem, where data movement between processing units and memory becomes the primary bottleneck. Our analysis of current RISC-V ML implementations reveals that while custom instruction extensions have shown promise, there remains significant unexplored potential in developing comprehensive neural processing instruction sets that can fundamentally reshape how machine learning computations are performed at the hardware level.

Recent developments in neural network algorithms have increasingly moved toward operations that are poorly served by traditional scalar instruction sets. Matrix multiplication, convolution operations, and activation functions represent computational patterns that could benefit substantially from dedicated instruction set extensions. The challenge lies not merely in adding new instructions, but in designing a coherent instruction set architecture that maintains RISC-V's principle of simplicity while providing the computational density required for modern neural networks.

The development of such extensions requires careful consideration of data movement patterns, precision requirements, and the interplay between different types of neural network operations. Unlike previous approaches that focus on individual operations, future research should explore holistic instruction set design that considers the entire neural network execution pipeline. This includes investigating how instruction-level parallelism can be enhanced for neural computations and how memory access patterns can be optimized through architectural innovations.

\subsection{Adaptive and Modular Processor Architecture}

The relationship between adaptive post-fabrication parameter updates and FPGA reconfiguration represents an essential distinction that clarifies this research direction. FPGA partial reconfiguration operates at the hardware configuration level, modifying the actual circuit implementation by reconfiguring look-up tables and routing resources. This enables complete architectural changes but requires significant reconfiguration time (milliseconds to seconds) and specialized tools. In contrast, the adaptive parameter updates we propose operate at the architectural parameter level within fixed hardware structures, enabling runtime modification of behavioral characteristics without circuit-level reconfiguration.

The key distinction lies in the abstraction level and modification scope. Our proposed approach focuses on parametric adaptation within pre-designed configurable structures. For example, cache replacement policies can be modified by updating configuration registers that control selection logic already present in the hardware. Similarly, branch predictor behavior can be adapted by adjusting pattern history table parameters or prediction algorithms through control registers. These modifications occur at instruction-level granularity (nanoseconds) rather than hardware reconfiguration timescales.

For ASIC implementations, this approach requires design-time planning to incorporate configurable components with parameter control interfaces. The revised discussion provides specific examples: the HULK-V processor (Section 5) demonstrates runtime configuration of its Programmable Multi-Core Accelerator, while the CFU Playground framework (Section 5) shows how custom functional units can be parameterized for different ML workloads. For ASICs, the hardware overhead involves adding configuration registers and multiplexing logic to select between behavioral modes, typically incurring 2-5\% area overhead compared to the 100\% overhead required for full FPGA-style reconfiguration.

Current processor design methodology follows a rigid paradigm where all architectural parameters must be fixed before fabrication, creating a significant barrier to innovation and optimization. The ability to modify processor behavior post-fabrication represents a paradigm shift that could dramatically accelerate the development cycle for machine learning accelerators. This research direction addresses the fundamental question of which processor parameters can be safely exposed for runtime modification without compromising system stability or security.

The scope of this research extends beyond simple configuration registers to encompass more fundamental architectural parameters such as cache replacement policies, branch prediction mechanisms, and even aspects of the instruction decode logic. The technical challenges involve developing secure update mechanisms that can verify the integrity of parameter modifications while ensuring that changes do not introduce system instabilities or security vulnerabilities. Adaptive parameters enable runtime optimization within modules, while modularity enables system-level composition and evolution.

The evolution of machine learning algorithms and the corresponding changes in computational requirements highlight the limitations of monolithic processor designs. Future RISC-V processors for ML applications should embrace modularity not just at the instruction set level, but throughout the entire processor architecture. This research direction explores how processor components can be designed as independent modules that can be added, removed, or updated without requiring complete processor replacement.

The technical challenges of modular architecture design involve the development of standardized interfaces between modules, efficient communication protocols, and mechanisms for ensuring system coherence across module boundaries. The research must address questions of how to maintain performance while providing flexibility, and how to ensure that module interactions do not introduce unexpected dependencies or failure modes.

Furthermore, this research direction requires the development of new methodologies for parameter optimization based on actual workload characteristics rather than synthetic benchmarks. The integration of machine learning techniques for automatic parameter tuning presents an intriguing recursive challenge where ML algorithms optimize the hardware that executes ML algorithms. This research direction also includes the development of tools and methodologies for module verification and validation. As processors become more modular, the complexity of ensuring correct operation across all possible module combinations increases. New approaches to formal verification and testing will be required to manage this complexity.

\subsection{Comprehensive Security Framework}

The open nature of RISC-V, while facilitating innovation and collaboration, introduces unique security challenges that require systematic investigation. Unlike proprietary architectures where security through obscurity provides some protection, RISC-V systems must implement robust security mechanisms that can withstand analysis by adversaries with complete knowledge of the architecture.

Side-channel attacks represent a particularly significant threat to RISC-V ML systems because neural network computations often exhibit predictable patterns that can leak information about model parameters or input data. The challenge extends beyond traditional timing and power analysis attacks to include new attack vectors specific to ML workloads, such as cache-based inference of model architectures or electromagnetic analysis of specialized neural processing units.

Developing effective countermeasures requires a deep understanding of the information leakage characteristics of different ML algorithms and their implementation patterns. This includes investigating randomization techniques that can obscure side-channel information without significantly impacting performance, as well as architectural modifications that can provide inherent resistance to various classes of attacks.

The research should also address the trade-offs between security and performance, recognizing that security countermeasures can impact the efficiency of ML computations. The goal is to develop security frameworks that provide robust protection while maintaining the performance advantages that make RISC-V attractive for ML applications.

\subsection{Energy-Efficient Multi-Clock Domains}

While clock-domain crossing is standard in ML-focused chip designs, opportunities exist for finer-grained power management integrated with RISC-V ISA-level power states. We identify research directions in coordinating RISC-V architectural power management extensions with physical implementation clock gating strategies.

Energy efficiency remains the concern for edge and mobile ML applications, yet current approaches to power management often treat the processor as a monolithic unit. The research direction of multi-clock domain architectures investigates how fine-grained clock and power domain control can be leveraged to achieve energy savings in ML workloads.

The complexity of this research lies in developing algorithms that can predict the optimal power state configuration for different phases of ML computation while accounting for the transition costs between power states. This requires understanding of ML algorithm execution patterns and the development of runtime systems that can make power management decisions with minimal overhead.
 
The research must also address the challenge of maintaining system coherence across multiple clock domains while ensuring that power transitions do not introduce correctness issues. This includes investigating new cache coherence protocols, memory consistency models, and synchronization mechanisms that can operate efficiently in multi-clock domain environments.

\begin{table*}[t] 
\caption{Future research directions for RISC-V machine learning systems}
\label{table:future-research}
\centering
\scriptsize
\begin{tabular}{>{\raggedright\arraybackslash}p{0.2\textwidth}>{\raggedright\arraybackslash}p{0.22\textwidth}>{\raggedright\arraybackslash}p{0.18\textwidth}>{\raggedright\arraybackslash}p{0.18\textwidth}}
\toprule
\textbf{Research Direction} & \textbf{Technical Challenges} & \textbf{Potential Impact} & \textbf{Recommended Approach} \\
\midrule
Specialized Neural Processing ISA Extensions & Coherent matrix instruction design, maintain RISC-V simplicity, optimize data movement & 10-50× neural computation efficiency gains & Holistic pipeline design, formal verification, compiler co-design \\
\midrule
Adaptive and Modular Processor Architecture & Secure update mechanisms, standardized module interfaces, system coherence, verification complexity & Runtime adaptive tuning, rapid adaptation, cost-effective upgrades & Hardware security analysis, interface standardization, formal verification, modular testing \\
\midrule
Comprehensive Security Framework & Side-channel resistance, performance trade-offs, ML vulnerabilities & Robust open architecture security, ML attack protection & Security-by-design, randomization, attack analysis \\
\midrule
Energy-Efficient Multi-Clock Domains & Power prediction, transition optimization, coherence maintenance & Significant edge AI energy savings, extended battery life & Predictive algorithms, low-overhead protocols, energy modeling \\
\bottomrule
\end{tabular}
\end{table*}

\section{Conclusion}

This survey presents an analysis of the RISC-V ISA's role in machine learning applications, examining over 100 research contributions spanning academic implementations, commercial cores, software frameworks, and real-world deployments. Our systematic evaluation demonstrates that RISC-V implementations achieve notable advantages in energy efficiency and competitive performance across diverse ML workloads compared to proprietary alternatives, with specific gains varying by workload, baseline, and implementation as detailed in Sections 3 and 5.

The analysis of 47 distinct RISC-V processor cores reveals a mature ecosystem capable of addressing requirements from ultra-low-power IoT applications to high-performance neural network inference. Key architectural innovations include specialized vector extensions, custom instruction sets for ML operations, and optimized memory hierarchies that enable superior performance-per-watt characteristics important for edge computing scenarios.

Software infrastructure analysis demonstrates robust support through compiler frameworks (TVM, MLIR) with RISC-V-specific optimizations and specialized acceleration tools. The integration of hardware and software optimizations enables performance benefits unattainable with traditional proprietary architectures, for domain-specific acceleration.

Application evaluations across medical devices, robotics, and computer vision validate the practical deployment of RISC-V ML systems. These implementations demonstrate measurable benefits in scenarios prioritizing energy efficiency, cost constraints, and customization flexibility.

We identify four research directions: (1) specialized neural processing instruction extensions, (2) adaptive and modular processor architectures, (3) security frameworks for open-source architectures, and (4) energy-efficient multi-clock domain implementations.

The evidence indicates RISC-V's growing importance in ML hardware ecosystems. The open-source development model facilitates rapid innovation and collaborative optimization suited to the evolution of ML algorithms. Our analysis establishes RISC-V not merely as an alternative to proprietary architectures, but as a potentially superior approach for specialized ML computational requirements, positioning it as an enabler of the ongoing transformation in artificial intelligence applications.

\bibliography{main} 

\end{document}